\documentclass{article}

    \PassOptionsToPackage{numbers, compress}{natbib}
\usepackage{natbib}

\usepackage[preprint]{neurips_2025}

\usepackage[utf8]{inputenc} % allow utf-8 input
\usepackage[T1]{fontenc}    % use 8-bit T1 fonts
\usepackage{hyperref}       % hyperlinks
\usepackage{url}            % simple URL typesetting
\usepackage{booktabs}       % professional-quality tables
\usepackage{svg}           % for including SVG images
\usepackage{amsfonts}       % blackboard math symbols
\usepackage{nicefrac}       % compact symbols for 1/2, etc.
\usepackage{microtype}      % microtypography
\usepackage{xcolor}         % colors
\usepackage{amsmath}

\usepackage{times}
\usepackage{latexsym}
\usepackage{booktabs}
\usepackage{multirow}
\usepackage{arydshln}
\usepackage{graphicx}
\usepackage{subcaption}
\usepackage{svg}
\usepackage{amssymb}
\usepackage{amsmath}
\usepackage{graphicx}
\usepackage{tcolorbox}
\tcbuselibrary{skins, breakable}
\usepackage{float}
\usepackage{caption}
\usepackage{amsmath}
\usepackage{amssymb}
\usepackage{enumitem}
\usepackage{lipsum}
\usepackage{hyperref}
\usepackage{colortbl}

 \usepackage{wrapfig}

\usepackage[T1]{fontenc}
\usepackage[utf8]{inputenc}

\usepackage{microtype}

\usepackage{inconsolata}

\usepackage{graphicx}

\title{Less is More Tokens: Efficient Math Reasoning via Difficulty-Aware Chain-of-Thought Distillation}

\author{
  Abdul Waheed$^{}$\thanks{Equal contribution. Laurie’s debugging efforts were generously supported by her avian companion (a bird), whose contributions we also warmly acknowledge.}\,
  ~~Chancharik Mitra$^{*}$\,
  ~~Laurie Z. Wang$^{*}$\,
  ~~Deva Ramanan\,
  ~~Bhiksha Raj \\
  Carnegie Mellon University \\
  \texttt{\normalsize \{abdulw,cmitra,zihaow3,deva,bhiksha\}@andrew.cmu.edu}
}

\begin{document}

\maketitle

\begin{abstract}
Chain-of-thought reasoning, while powerful, can produce unnecessarily verbose output for simpler problems. We present a framework for difficulty-aware reasoning that teaches models to dynamically adjust reasoning depth based on problem complexity. Remarkably, we show that models can be endowed with such dynamic inference pathways without any architectural modifications; we simply post-train on data that is carefully curated to include chain-of-thought traces that are proportional in length to problem difficulty.  
Our analysis reveals that post-training via supervised fine-tuning (SFT) primarily captures patterns like reasoning length and format, while direct preference optimization (DPO) preserves reasoning accuracy, with their combination reducing length and maintaining or improving performance. Both quantitative metrics and qualitative assessments confirm that models can learn to "think proportionally"—reasoning minimally on simple problems while maintaining depth for complex ones.
\end{abstract}

% Introduction
\section{Introduction}\label{sec:introduction}

Chain-of-thought (CoT) prompting~\citep{Wei2022ChainOT} has emerged as a powerful paradigm to improve the reasoning abilities of large language models (LLMs) across domains such as math, coding, and commonsense reasoning~\cite{chuCoTReasoningSurvey2024, patil2025advancingreasoninglargelanguage, deepseek2025r1, openai2024openaio1card}. By generating intermediate steps, CoT enables models to perform complex multi-step inferences more accurately. However, a critical limitation remains: these reasoning traces are often overly verbose, even for problems that require little or no intermediate reasoning. This leads to inefficient "over-thinking" during inference —longer outputs that result in increased latency and computational cost~\cite{chen2025think23overthinkingo1like}. Indeed, recent estimates of energy costs suggest that inference during model deployment will dwarf the cost and environmental impact of training~\cite{samsi2023words}.

% In real-world settings, not all problems require the same amount of reasoning. For instance, simple arithmetic questions can be solved in a single step, while harder algebraic or geometric problems demand detailed multi-step explanations. Yet most CoT-enabled models apply a fixed reasoning strategy across all inputs. Recent efforts such as L1~\cite{Aggarwal2023LCPO}, introduce explicit length control through reinforcement learning, while work like “Learning How Hard to Think”~\cite{damani2024learninghardthinkinputadaptive} adapts compute allocation dynamically at inference time. However, these methods either enforce a fixed budget per query or modulate external procedures like a router or sampling, without teaching the model to internally vary its reasoning length. In our work, we address this problem by training the models to \textit{dynamically adjust the granularity and length of their reasoning trajectories based on the difficulty of the input problem}.

In real-world settings, not all problems require the same amount of reasoning. For instance, simple arithmetic questions can be solved in a single step, while harder algebraic or geometric problems demand detailed multi-step explanations. Yet most CoT-enabled models apply a fixed reasoning strategy across all inputs. This is misaligned with how humans allocate cognitive effort: we intuitively adjust how much we think based on task complexity.

% \begin{quote}
% \textit{“System 1 operates automatically and quickly, with little or no effort and no sense of voluntary control. System 2 allocates attention to the effortful mental activities that demand it, including complex computations.”}  
% \hfill--- Daniel Kahneman, \textit{Thinking, Fast and Slow}~\cite{kahneman2011thinking}
% \end{quote}

Recent efforts, such as L1~\cite{Aggarwal2023LCPO}, introduce explicit length control through reinforcement learning, while work like “Learning How Hard to Think”~\cite{damani2024learninghardthinkinputadaptive} adapts compute allocation dynamically at inference time. However, these methods either enforce a fixed budget per query or modulate external procedures like a router or sampling, without teaching the model to internally vary its reasoning length. In our work, we address this problem by training the models to \textit{dynamically adjust the granularity and length of their reasoning trajectories based on the difficulty of the input problem}.

We propose a general framework for difficulty-aware chain-of-thought reasoning. First, we estimate the difficulty of each math problem using a LLM based grader aligned with AoPS difficulty scales~\footnote{\url{https://artofproblemsolving.com/wiki/index.php/AoPS_Wiki:Competition_ratings}}. Next, we compress long CoT traces generated by a strong LLM into concise reasoning sequences, where the amount of compression is guided by the problem’s difficulty. This produces supervision that reflects the ideal level of reasoning granularity per input. We then fine-tune student models using both supervised fine-tuning (SFT) and direct preference optimization (DPO)~\cite{rafailov2024directpreferenceoptimizationlanguage}, training them to imitate difficulty-aligned traces—thinking minimally on easy questions and extensively on hard ones.

We empirically study this setup across unimodal and multimodal models for the math reasoning task. Our work centers around two key research questions: \textbf{(RQ1)} Can difficulty-adaptive models match or exceed the accuracy of full CoT models while reducing unnecessary reasoning? \textbf{(RQ2)} Do they offer better efficiency–accuracy trade-offs than static or explicitly length-controlled baselines? Our results demonstrate that difficulty-aware reasoning can reduce reasoning token usage by up to 30\%, with minimum to no loss—and in some cases, gains—in final answer accuracy. Compared to uniform-length CoT or fixed-budget approaches, our models learn to allocate reasoning effort more effectively across problem types.
% DEVA: I'd state "difficulty-aware reasoning can reduce reasoning token usage by up to 10X for simple problems (and 30% on average)"

In summary, we present a systematic study of difficulty-aware reasoning in math problem solving. We introduce a scalable framework for generating and training on difficulty-aligned CoT traces, and show that LLMs can be taught to reason proportionally to the complexity of the task. Our findings open up promising directions for building more efficient, interpretable, and adaptive reasoning agents—models that not only solve problems, but also know how hard to think.

% Preliminaries
\section{Preliminaries}\label{sec:preliminaries}

Large Language Models (LLMs) have shown strong capabilities in multi-step reasoning through \textit{Chain-of-Thought} (CoT) prompting. In this paradigm, models produce explicit intermediate steps when solving a problem, which helps improve both interpretability and answer accuracy. However, CoT traces—especially those generated by powerful reasoning or what we call ``thinking'' models—are often excessively verbose~\citep{chen2025think23overthinkingo1like}. They frequently contain \textit{trivial} and/or \textit{redundant} reasoning steps, even for simple problems such as arithmetic or elementary algebra. This verbosity inflates the number of generated tokens, thereby increasing inference latency and computational cost—a key limitation for real-time or resource-constrained deployment.

A central insight motivating our work is that the complexity of reasoning required to solve a problem should vary with the intrinsic difficulty of that problem. Yet, existing CoT methods typically produce reasoning traces of uniform verbosity across a wide range of problem difficulties. This motivates our core hypothesis: \textit{the length and granularity of CoT reasoning should scale with the complexity of the problem}.

Let \( x \in \mathcal{X} \) denote an input math problem, and let \( y \in \mathcal{Y} \) be the corresponding correct final answer. A CoT-enabled model \( f \) solves the problem by generating a reasoning trace \( r \in \mathcal{R} \) and a final answer \( y \), such that
\[
f(x) = (r, y), \quad r = (r_1, r_2, \dots, r_T),
\]
where \( T = |r| \) is the length (in tokens) of the reasoning trace. In current models, the expected reasoning length \( \mathbb{E}[T \mid d(x)] \) tends to be invariant to the problem difficulty \( d(x) \in [1, 10] \), i.e.,
\[
\mathbb{E}[T \mid d(x)] \approx \text{constant}, \quad \forall x \in \mathcal{X}.
\]

This lack of adaptability leads to inefficient reasoning. Our goal is to design models where the expected reasoning length is positively correlated with the problem difficulty:
\[
\mathbb{E}[T \mid d(x)] \propto d(x).
\]
Such models would produce concise reasoning for easy problems (\( d(x) \ll 10 \)) and more elaborate reasoning for difficult ones (\( d(x) \approx 10 \)), allocating computational effort proportionally to task complexity.

To enable this, we adopt a teacher–student distillation framework. A high-capacity teacher model \( f_T \) generates detailed reasoning traces \( r_T \), which are then passed through a difficulty-aware summarization function:
\[
s(r_T, d(x)) \rightarrow \tilde{r},
\]
where \( \tilde{r} \) is a compressed reasoning trace that reflects the difficulty \( d(x) \). A student model \( f_S \) is trained to map
\[
x \mapsto (\tilde{r}, y), \quad \text{with } |\tilde{r}| \approx \alpha \cdot d(x), \quad \alpha > 0,
\]
thus learning to internalize difficulty-aware reasoning behavior.

This formulation serves two objectives: (1) \textit{Efficiency}, by reducing average token generation per sample, and (2) \textit{Adaptivity}, by tailoring the reasoning depth to the task at hand. Our evaluations assess not only answer correctness but also how well the model modulates reasoning length in accordance with \( d(x) \), providing a joint measure of performance and computational efficiency.

% Method
\section{Method}\label{sec:method}
We present a general framework for difficulty-aware chain-of-thought (CoT) distillation that enables models to adapt their reasoning verbosity based on problem difficulty. Our approach consists of two main stages: (1) data generation through difficulty-aligned compression of \textit{overly} verbose reasoning traces for both unimodal and multimodal mathematical problems, and (2) training using a combination of supervised fine-tuning (SFT) and direct preference optimization (DPO). The key intuition is that by teaching models to imitate traces that are adapted to the complexity of each input, we can encourage efficient and interpretable reasoning without compromising accuracy.

\subsection{Data Generation}

We begin with a collection of math problems \( \{x_i\}_{i=1}^N \), each associated with a long, detailed reasoning trace \( r_i^{\text{long}} \) generated by a high-capacity teacher model. While these traces are generally accurate, they tend to be overly verbose—offering exhaustive reasoning even for trivial problems. To correct this mismatch, we design a pipeline that compresses each trace into a more concise form based on the difficulty of the underlying problem. To do this, we use \textit{GTP-4o-mini} (temperature = 0.6) with the prompt shown in Figure~\ref{fig:unimodal_compression_prompt} and~\ref{fig:multimodal_compression_prompt}.

\paragraph{Difficulty Estimation.} To assign a scalar difficulty score \( d(x_i) \in [1, 10] \) to each problem, we use \texttt{GPT-4o-mini} (temperature=0). The prompt, as shown in Figure~\ref{fig:difficulty_ratings} is designed to align with the Art of Problem Solving (AoPS) competition difficulty ratings, following the rubric used in SkyThought~\cite{sky_t1_2025}. This procedure ensures that estimated difficulty scores are stable, human-aligned, and reflect the true complexity of the reasoning required (see Appendix~\ref{appsubsubsec:rating}).

\paragraph{Difficulty-Aware Compression.} Given the difficulty score \( d(x_i) \), we compress the long CoT trace \( r_i^{\text{long}} \) into a shorter version \( \tilde{r}_i \) using a prompt-based summarization function:
\[
\tilde{r}_i = s(r_i^{\text{long}}, d(x_i)),
\]
where \( s(\cdot, \cdot) \) represents a deterministic compression process instantiated via few-shot prompting. This function adapts the verbosity of the reasoning proportionally to the problem's difficulty—minimal steps for easy problems and more elaborate chains for hard ones—while preserving logical correctness and final answer. We generate the resulting training tuples \( (x_i, \tilde{r}_i, y_i) \), where \( y_i \) is the correct final answer.

In Table~\ref{tab:compression_ratio}, we report the compression ratios across different difficulty levels for both unimodal and multimodal settings. In the unimodal setup, we observe that harder problems tend to require longer reasoning chains, leading to higher compression ratios as difficulty increases. In contrast, the multimodal setup exhibits relatively uniform CoT lengths across difficulty levels, suggesting that the initial reasoning traces are already more optimized. As a result, the compression ratios remain constant or even decrease with increasing difficulty.

%DEVA: Need to define the compression ratio in Table 1 with a statement like "Difficulty 1 problems can be compressed to 79% of the raw chain-of-thought length. Why does the multimodal CR *decrease* for increased difficulty?

% Added explanation

\paragraph{Unimodal Data.} For unimodal reasoning, we curate approximately 125K unique problems from the \emph{CoT} split of OpenMathReasoning~\cite{moshkov2025aimo2}, with long reasoning traces generated using DeepSeek-R1~\cite{deepseekai2025deepseekr1incentivizingreasoningcapability} and Qwen-QwQ-32B~\cite{qwq32b, qwen2.5}. The problems span diverse topics (algebra, geometry, number theory) and originate from AoPS forums\footnote{\url{https://artofproblemsolving.com/community}}.

\paragraph{Multimodal Data.} For multimodal data, we use subset of LLaVA-CoT-100K~\citep{xu2024llavacotletvisionlanguage}, comprising approximately 60{,}000 examples. Each example includes an extended chain‐of‐thought generated by \textbf{GPT-4o}, which we then condense using \textbf{GPT-4o-mini} to produce concise summaries. These summarized reasoning trajectories serve as the training data for our multimodal models.

\subsection{Training}
% \begin{wraptable}[18]{r}{0.5\textwidth} % r = right side, width = 0.5\textwidth
% \centering
% \setlength{\tabcolsep}{3pt}
% \renewcommand{\arraystretch}{0.8}
% \resizebox{\linewidth}{!}{%
% \begin{tabular}{ccc}
% \toprule
% Difficulty & CR (Unimodal) & CR (Multimodal) \\
% \midrule
% 1          & 79.1          & 49.9            \\
% 2          & 84.5          & 47.6            \\
% 3          & 87.0          & 43.6            \\
% 4          & 88.0          & 40.6            \\
% 5          & 88.8          & 35.1            \\
% 6          & 89.4          & 29.7            \\
% 7          & 90.5          & 16.9            \\
% 8          & 90.6          & NA              \\
% 9          & 90.4          & NA              \\
% 10         & 88.8          & NA              \\
% \bottomrule
% \end{tabular}
% }
% \caption{Compression ratios (\%) across difficulty levels for unimodal and multimodal tasks. Higher values indicate greater reduction in reasoning length. Unimodal compression increases with difficulty, while multimodal ratios remain flat or decrease.}
% \label{tab:compression_ratio}
% \end{wraptable}

\begin{wraptable}[16]{r}{0.5\textwidth}
\vspace{-0.5em} % move table up slightly
\centering
\setlength{\tabcolsep}{3pt}
\renewcommand{\arraystretch}{0.8}
\resizebox{\linewidth}{!}{%
\begin{tabular}{ccc}
\toprule
Difficulty & CR (Unimodal) & CR (Multimodal) \\
\midrule
1 & 79.1 & 49.9 \\
2 & 84.5 & 47.6 \\
3 & 87.0 & 43.6 \\
4 & 88.0 & 40.6 \\
5 & 88.8 & 35.1 \\
6 & 89.4 & 29.7 \\
7 & 90.5 & 16.9 \\
8 & 90.6 & NA \\
9 & 90.4 & NA \\
10 & 88.8 & NA \\
\bottomrule
\end{tabular}}
% \vspace{0.5em} % prevent collision with equation
\caption{Compression ratios (\%) across difficulty levels for unimodal and multimodal tasks. Higher values indicate greater reduction in reasoning length. Unimodal compression rises with difficulty, while multimodal ratios remain flat or decrease.}
\label{tab:compression_ratio}
\end{wraptable}

Once the difficulty-aligned training data is generated, we train compact student models to replicate this behavior using both supervised and preference-based learning paradigms. We experiment with Qwen2.5-Math (1.5B, 7B) for unimodal, and Qwen2.5-VL (3B, 7B-Instruct) for multimodal setting.

\paragraph{Supervised Fine-Tuning (SFT).}
SFT serves as the first stage of training, where the model learns to map each input problem to its compressed trace:
\[
\min_{\theta} \; \mathcal{L}_{\text{SFT}} = \sum_{i=1}^{N} \text{CE}\left(f_\theta(x_i), \tilde{r}_i\right),
\]
where \( f_\theta \) is the student model and CE denotes the cross-entropy loss between its output and the compressed trace. This objective encourages the model to internalize reasoning styles that are aligned with the problem’s complexity.

\paragraph{Direct Preference Optimization (DPO).}
To further refine reasoning behavior, we apply DPO~\cite{rafailov2024directpreferenceoptimizationlanguage} using preference data. For each input, we treat the compressed trace \( \tilde{r}_i \) as the preferred response and the original long trace \( r_i^{\text{long}} \) as the rejected one. The training objective is:
\[
\resizebox{\columnwidth}{!}{$
\min_{\theta} \; \mathcal{L}_{\text{DPO}} = -\sum_{i=1}^{N} \log \frac{\exp\left(\beta \cdot \text{KL}(f_\theta(x_i), \tilde{r}_i)\right)}{\exp\left(\beta \cdot \text{KL}(f_\theta(x_i), \tilde{r}_i)\right) + \exp\left(\beta \cdot \text{KL}(f_\theta(x_i), r_i^{\text{long}})\right)},
$}
\]
where \( \beta \) is a temperature parameter and KL denotes the Kullback–Leibler divergence. This step explicitly teaches the model to prefer compressed, efficient reasoning over redundant verbosity, guided by human-aligned preferences.

\paragraph{Hybrid Training (SFT + DPO).}
We also explore a hybrid training setup where SFT is followed by DPO. This curriculum allows the model to first acquire difficulty-conditioned reasoning patterns through imitation (SFT), and then refine them via preference-based contrastive learning (DPO). We find that this combination yields the most consistent gains across our benchmarks (see Section~\ref{sec:results}).

Our method systematically generates and distills difficulty-aware CoT traces, encouraging LLMs to “think harder” only when needed. The approach generalizes across modalities and model sizes, and supports efficient reasoning without sacrificing final answer accuracy. In the next section, we describe our experimental setup and metrics for evaluating reasoning efficiency and correctness.

% Experiments
\section{Experimental Setup}\label{sec:experiments}

In this section, we detail the experimental setup used to evaluate our difficulty-aware chain-of-thought distillation framework. Our goal is to assess both the reasoning performance and token efficiency of models trained with difficulty-aligned CoT traces. We evaluate our setup across a diverse suite of unimodal and multimodal benchmarks, compare against strong reasoning and instruction-tuned baselines, and analyze model outputs across problem difficulty levels. 

\subsection{Models and Datasets} 
We train our models using the difficulty-aware datasets generated as described in Section~\ref{sec:method}. For \textit{unimodal} evaluation, we use five math-focused reasoning benchmarks: AIME 2025, MATH~\cite{hendrycks2021measuringmathematicalproblemsolving}, AMC, GSM8K~\cite{cobbe2021gsm8k}, and HMMT~\cite{balunovic_srimatharena_2025} (2024 \& 2025). For \textit{multimodal} evaluation, we use five widely adopted visual reasoning benchmarks: MathVerse~\cite{zhang2024mathversedoesmultimodalllm}, HLE (V)~\cite{phan2025humanitysexam}, MathVision~\cite{wang2024measuring}, OlympiadBench (V)~\cite{he2024olympiadbenchchallengingbenchmarkpromoting}, MMStar~\cite{chen2024rightwayevaluatinglarge}. In all multimodal benchmarks, we retain only single-image samples with free-form answer generation, excluding multiple-image inputs and multiple-choice questions when not entirely supported.

\subsection{Baselines} 
For \textit{unimodal} models, we compare against two types of baselines: (1) models trained on long chain-of-thought traces, such as the Qwen series fine-tuned on DeepSeek-R1-generated reasoning; and (2) explicit length-controlled reasoning models like L1~\cite{aggarwal2025l1controllinglongreasoning}. For \textit{multimodal} models, we compare against: (1) base instruction-tuned models such as Qwen2.5-VL-3B and Qwen2.5-VL-7B-Instruct; and (2) multimodal reasoning models such as LLaVA-CoT-11B~\cite{xu2024llavacotletvisionlanguage}. These serve as both upper and lower bounds in terms of token budget.

\subsection{Evaluation}
Our evaluation consists of three key steps: (1) estimating the difficulty of each evaluation problem using the AoPS-based rating system; (2) \textit{pass@1} rate across benchmarks; and (3) calculating the average number of tokens used across different evaluation benchmarks. For unimodal we directly extract the and match it with the ground truth answer; however, for multimodal models, we observe occasional failures to adhere to prompt formatting, leading to difficulties in automatic answer extraction. To address this, we use a verifier model: given the model-generated solution and the ground-truth answer, we prompt \texttt{GPT-4o-mini} (temperature = 0) to determine correctness. The verification prompt is shown in Appendix~\ref{appsubsubsec:verification}, Figure~\ref{fig:verification_prompt}.

\paragraph{LLM-as-Judge Evaluation}
Prior work~\cite{boye2025largelanguagemodelsmathematical} shows that incorrect reasoning can lead to correct answers, while even correct reasoning can sometimes produce the wrong final answer. Further, a key subgoal of our work is to improve the readability of reasoning traces by compressing them, removing redundant or overly verbose content. However, there are no existing automatic metrics for evaluating this aspect. To address this, we perform a human-aligned evaluation of reasoning traces generated by five different models on 100 math problems using the LLM-as-Judge~\cite{zheng2023judgingllmasajudgemtbenchchatbot, gu2025surveyllmasajudge}. Each trace is rated across four dimensions: \textit{clarity}, \textit{completeness}, \textit{correctness}, and \textit{redundancy}, using a 1–5 Likert scale. We design clear annotation guidelines for what constitutes a rating of 1 versus 5 for each criterion. These ratings are then assigned by prompting \emph{GPT-4o-mini} (temperature=0).

\subsection{Hyperparameters and Implementation Details}
We train all models for three epochs with a maximum sequence length of 4096 tokens. Training is conducted using the LLaMA-Factory framework~\cite{zheng2024llamafactory}. The hyperparameters are provided in Table~\ref{tab:hyperparams} in Appendix~\ref{appsubsec:hyperparams}. For evaluation, we use \texttt{VLMEvalKit}~\cite{duan2024vlmevalkit} for multimodal models and Hugging Face Transformers~\cite{wolf-etal-2020-transformers} for text-only models. All models are trained using mixed precision (fp16) with greedy decoding. Training is carried out on 8×H100 GPUs, and evaluations are run on clusters with varying configurations including 2×H100 and 4×A100 setups. No extensive hyperparameter search is performed due to resource constraints.

Our experimental setup enables a comprehensive evaluation of both accuracy and reasoning efficiency. By comparing models across a wide range of math benchmarks—spanning symbolic and visual reasoning—we demonstrate the impact of difficulty-aware supervision on reasoning behavior. Our results, presented in the next section, reveal clear trends in token-efficiency and reasoning alignment, highlighting the strengths and trade-offs of SFT, DPO, and their combination under our proposed framework. A detailed list of hyperparameters is provided in Appendix~\ref{appsubsec:hyperparams}, Table~\ref{tab:hyperparams}.

% Results
\section{Results}\label{sec:results}

% Section header for qualitative evaluation

% DEVA: Personally, I would move some of this content to the captions. My mental model of time strapped reviewer / area chair is someone that just reads the intro + figures-with-captions, but none of the figures or tables have self-contained captions. At the very least, I'd explain the A(B) format of Table 2+3 is A = pass@1 and B = token count. I still think it would help to show the token compression ratio for easy test problems, which seems to be 10X (according to Fig1-a which somes a similar stat but for train, a bigger isue is that the multimodal results are strange).

% Yes -- all captions will be updated. As of now, they're a placeholder.
% Added explanation -- why trend in multimodal trend is opposite. 

We present empirical results demonstrating the effectiveness of our difficulty-aware chain-of-thought (CoT) compression framework across both unimodal (text-only) and multimodal (vision-language) setups. Our evaluation begins by assigning difficulty ratings to each problem using the method described in Section~\ref{sec:method}, applied to both training and evaluation datasets. As shown in Figure~\ref{fig:difficulty_ratings}, text-only math problems are generally rated as more difficult than multimodal ones based on the AoPS difficulty scale. We then compress the original long CoT traces into concise, non-redundant versions, guided by these difficulty scores. We compare several prompting strategies for this compression and report ablation results in Appendix~\ref{appsubsec:results}. 

After compressing the long chain-of-thought reasoning, we train and evaluate multiple student models—text-only models (1.5B and 7B) and multimodal models (3B and 7B)—on the compressed traces to assess the effectiveness of our approach. We evaluate the trained models using \emph{Pass@1} and average token count across a wide range of standard evaluation setups. We use `GPT-4o'~\footnote{\url{https://github.com/openai/tiktoken}} to calculate token count across our experiments. 

\paragraph{Unimodal}
We evaluate our difficulty-aware compression approach using two text-only student models: R1-Distill-Qwen-1.5B and R1-Distill-Qwen-7B. Combining supervised fine-tuning (SFT) with direct preference optimization (DPO) consistently outperforms either method alone. For instance, on AIME, the 1.5B model improves from 5.0 (SFT) and 18.4 (DPO) to 23.4 with SFT+DPO, while the 7B variant achieves 31.7, up from just 8.4 with SFT alone.

\begin{table*}[ht!]
\centering
\footnotesize % Use footnotesize for compact representation
\setlength{\tabcolsep}{5pt} 
\renewcommand{\arraystretch}{1.0}  
\begin{tabular}{@{}lcccccc@{}}
\toprule
\textbf{Setup} & \textbf{AIME} & \textbf{AMC} & \textbf{Math} & \textbf{GSM8K} & \textbf{HMMT} \\
\midrule
\multicolumn{6}{@{}l}{\textit{Base}} \\[0.5ex]
\quad R1-Distill-Qwen-1.5B & 20.0 (6768) & 54.2 (5381) & 77.0 (3588) & 83.9 (1660) & 12.2 (7176) \\
\quad R1-Distill-Qwen-7B   & 35.0 (6360) & 71.1 (4723) & 86.0 (3050) & 88.0 (553) & 24.4 (6876) \\[0.5ex]
\midrule
\multicolumn{6}{@{}l}{\textit{Thinking}} \\[0.5ex]
\quad L1-Qwen-1.5B-Max    & 24.4 (2280) & 67.5 (1976) & 83.6 (1693) & 87.3 (1479) & 11.1 (2272) \\
\quad Nemotron-1.5B     & 33.3 (6692) & 62.7 (5186) & 83.0 (3419) & 75.9 (2965) & 14.4 (7751) \\
\quad Nemotron-7B       & 31.7 (6770) & 62.7 (5196) & 86.2 (3112) & 89.7 (1850) & 28.9 (6969) \\[0.5ex]
\midrule
\multicolumn{6}{@{}l}{\textit{R1-Distill-Qwen-1.5B}} \\[0.5ex]
\quad SFT      & 5.0 (1081) & 38.6 (699) & 65.2 (526) & 77.6 (241) & 0.0 (845) \\
\quad DPO      & 18.4 (6964) & 50.6 (5524) & 74.6 (3864) & 82.9 (1733) & 10.0 (7388) \\
\quad SFT + DPO & 23.4 (6192) & 57.8 (4893) & 79.4 (3269) & 82.3 (1527) & 12.2 (6458) \\[0.5ex]
\midrule
\multicolumn{6}{@{}l}{\textit{R1-Distill-Qwen-7B}} \\[0.5ex]
\quad SFT      & 8.4 (864) & 51.8 (789) & 75.4 (474) & 88.5 (244) & 2.2 (808) \\
\quad DPO      & 25.0 (6584) & 62.7 (4932) & 86.0 (3062) & 89.2 (592) & 17.8 (6940) \\
\quad SFT + DPO & 31.7 (5724) & 69.9 (4251) & 88.8 (2745) & 86.9 (498) & 20.0 (6188) \\
\bottomrule
\end{tabular}
\caption{Performance of unimodal models across seven math reasoning benchmarks. Each cell reports \textbf{Pass@1} with \textbf{average token count} in parentheses. Our proposed SFT+DPO models are on par or outperform base and individually fine-tuned variants across most benchmarks while generating shorter reasoning traces.}
\label{tab:unimodal_results}
\end{table*}

More importantly, our approach yields substantial token efficiency gains. The 1.5B SFT+DPO model surpasses its base counterpart on AIME (23.4 vs. 20.0) while using 8.5\% fewer tokens (6,192 vs. 6,768). The 7B variant maintains comparable performance (31.7 vs. 35.0 base) with 10\% token reduction (5,724 vs. 6,360). These efficiency gains are particularly pronounced on certain benchmarks—on GSM8K, the 1.5B model achieves near-base accuracy (82.3 vs. 83.9) using only 1,527 tokens compared to 1,660 for the base model. Overall, difficulty-aware CoT compression with preference optimization reduces reasoning trace length by approximately 8–10\% while maintaining or improving accuracy across most benchmarks.

% \paragraph{Unimodal}
% We evaluate our difficulty-aware compression approach using two text-only student models: R1-Distill-Qwen-1.5B and R1-Distill-Qwen-7B. Across all benchmarks, combining supervised fine-tuning (SFT) with direct preference optimization (DPO) gives the strongest performance. For instance, on the AIME benchmark, R1-Distill-Qwen-1.5B improves from 5.0 (SFT) and 18.4 (DPO) to 23.4 with SFT+DPO. Similarly, the 7B variant achieves 31.7 on AIME with SFT+DPO, compared to just 8.4 with SFT alone. This performance gain is consistent across benchmarks such as AMC, Math, and JEEBench.

% Crucially, our difficulty-aware compression yields substantial gains in token efficiency. For example, compared to the base R1-Distill-Qwen-7B model which uses 6,360 tokens on AIME, the SFT+DPO version achieves better accuracy (31.7 vs. 35.0) while reducing the average token count to 6,308—a marginal drop. However, the most pronounced improvements are seen in smaller models: the 1.5B SFT+DPO model achieves 23.4 on AIME using just 6,432 tokens, compared to the base model's 6,768 tokens—a 5\% reduction in tokens. On GSM8K, we observe even greater efficiency, where the 1.5B model with SFT+DPO achieves 82.3 accuracy using only 1,097 tokens—a 34\% reduction from the base model's 1,660 tokens.

% These results show that difficulty-aware CoT compression, when combined with preference optimization, can significantly reduce reasoning trace length, achieving up to 30–35\% token savings—without sacrificing accuracy.

\begin{figure*}[h!]
    \centering
    \includegraphics[width=\linewidth]{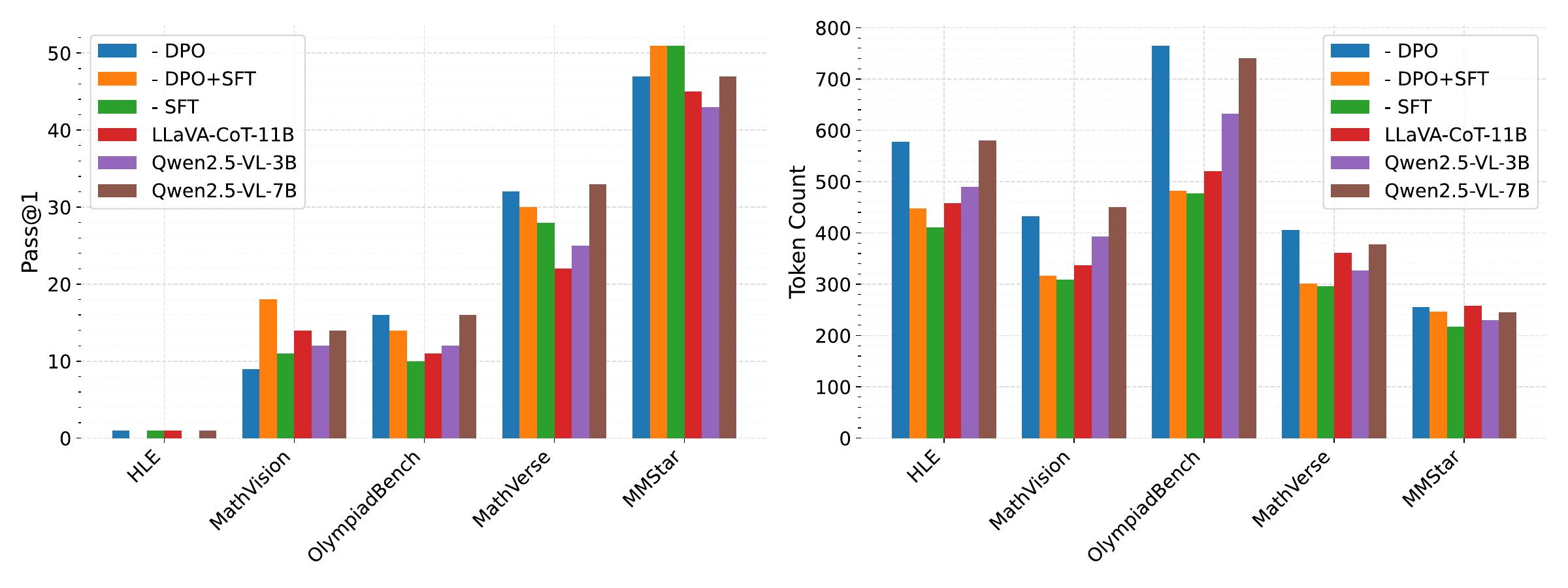}
    \caption{The left bar plot shows pass@1 and the right plot shows average token count in reasoning outputs across multimodal benchmarks. SFT, DPO, and SFT+DPO are our models trained on Qwen-2.5-VL-7B-Instruct base.}
    \label{fig:mm_pass@1_and_token_count}
\end{figure*}

% \input{tables/multimodal_results}
\paragraph{Multimodal}  
In our main results (Figure~\ref{fig:mm_pass@1_and_token_count}), we report \textit{pass@1} with a verifier alongside the average reasoning token counts across multimodal math benchmarks. We find that supervised fine-tuning (SFT) alone already provides strong performance and efficiency: for example, the 7B SFT model attains 28\% on MathVerse, outperforming both LLaVA-CoT-11B (22\%) and Qwen2.5-VL-3B (25\%), and achieves 51\% on MMStar compared to 45\% and 43\% from the same baselines. Direct preference optimization (DPO) alone boosts performance selectively—for instance, reaching 16\% on OlympiadBench, on par with Qwen2.5-VL-7B, though it lags behind SFT on MathVision (9\% vs.\ 14\%). The combined SFT+DPO consistently delivers the strongest overall outcomes, such as 18\% on MathVision (vs.\ 14\% for Qwen2.5-VL-7B) and 51\% on MMStar, highlighting the complementarity of the two approaches.  

From Figure~\ref{fig:mm_pass@1_and_token_count} (right), we observe that SFT also confers notable efficiency gains, reducing token usage by 25--30\% relative to its 7B teacher. For example, average token count drops from 580 to 411 on HLE and from 378 to 296 on MathVerse while maintaining comparable or superior accuracy. DPO by itself generally increases reasoning length, indicating a trade-off between alignment and efficiency. Importantly, the combined SFT+DPO model retains SFT’s efficiency while improving accuracy: on MathVision, it produces 317 tokens on average (vs.\ 308 for SFT) while boosting \textit{pass@1} from 14\% to 18\%. Overall, the 7B SFT+DPO model reduces token usage by more than 10\% compared to the teacher while achieving +4--5 points higher accuracy across benchmarks.  

These results demonstrate that our approach not only achieves competitive or superior accuracy relative to stronger baselines but also maintains compact reasoning traces. In practice, this translates to reductions of up to 30--35\% in reasoning length, while allocating tokens adaptively across problem difficulty. Unlike LLaVA-CoT, which shows limited sensitivity to task difficulty, our SFT+DPO models expend more tokens on challenging tasks while compressing easy cases, yielding a balanced trade-off between accuracy and efficiency.

\noindent\textbf{SFT Learns Length and Format.} Our initial experiments with supervised fine-tuning (SFT) revealed that while models effectively learned to mimic the structure and brevity of our difficulty-aware summarized traces, this came at the cost of reduced accuracy. As seen in Table~\ref{tab:unimodal_results} and Figure~\ref{fig:mm_pass@1_and_token_count}, SFT models produced significantly shorter outputs aligned with our compression objectives and consistently adopted the clean, numbered reasoning format used by the \texttt{gpt4o\_DA2} summarizer (see Appendix~\ref{appsubsec:results}). This indicates that SFT is effective at learning stylistic and structural patterns present in the training data. However, this stylistic conformity did not translate into better task performance. Despite generating compact and well-formatted responses, SFT models often failed to arrive at the correct final answer, underperforming relative to their base counterparts. 

% These findings suggest that while SFT captures surface-level features well, it struggles to maintain the depth of reasoning required for high accuracy—especially when trained only on compressed traces. This insight led us to explore preference-based methods like DPO, which aim to directly optimize for reasoning quality while retaining the benefits of concise output.

\noindent\textbf{DPO Learns Reasoning Accuracy.} In contrast to our SFT results, Direct Preference Optimization (DPO) demonstrated strong improvements in reasoning accuracy across both unimodal and multimodal settings. As shown in Table~\ref{tab:unimodal_results} and Figure~\ref{fig:mm_pass@1_and_token_count}, DPO-trained models consistently matched or outperformed their base counterparts in accuracy, confirming its effectiveness in aligning model outputs with correctness-oriented preferences. However, DPO offered limited gains in token efficiency. Unlike SFT, which significantly reduced reasoning length, DPO models retained much of the original verbosity, with only modest compression observed. Qualitative analysis (Appendix~\ref{appsubsec:results}) supports the observation that DPO outputs emphasized completeness and correctness over brevity.

% , often resulting in more detailed, less structured chains of thought. These results highlight a trade-off—DPO preserves reasoning depth but lacks the compactness achieved through SFT—motivating our next step: combining both methods to balance accuracy and conciseness.

\noindent\textbf{SFT+DPO Balances Length and Correctness Objectives.}
Combining supervised fine-tuning (SFT) with direct preference optimization (DPO) provides an effective strategy for balancing reasoning conciseness with accuracy. As shown in Tables~\ref{tab:unimodal_results} and Figure~\ref{fig:mm_pass@1_and_token_count}, models trained with this two-stage approach achieve up to 30\% reduction in token usage while maintaining or improving final answer accuracy across unimodal and multimodal benchmarks. SFT encourages the model to mimic surface-level patterns in the compressed traces—such as structure and brevity—by minimizing cross-entropy loss. However, it does not explicitly optimize for correctness, especially when critical reasoning steps are condensed in the summaries.

DPO complements this by learning from pairwise preferences, optimizing for correctness and clarity without relying on exact sequence matching. When preferred responses are both accurate and well-formatted, DPO implicitly promotes efficient reasoning while preserving answer quality. Additionally, the KL-regularization term in DPO stabilizes performance, preventing overfitting or degradation in reasoning. When applied after SFT, DPO refines correctness without sacrificing conciseness, resulting in models that produce structured, accurate, and difficulty-aware reasoning traces—directly addressing our core research question (RQ1).

\paragraph{Qualitative Evaluation with LLM-as-Judge}
We conduct a human-aligned evaluation using the LLM-as-Judge framework, where \texttt{GPT-4o-mini} scores reasoning traces on four dimensions: \textit{clarity}, \textit{completeness}, \textit{correctness}, and \textit{redundancy}.
\begin{figure*}[ht!]
    \centering
    \begin{subfigure}[b]{0.5\textwidth}
        \includegraphics[width=\linewidth]{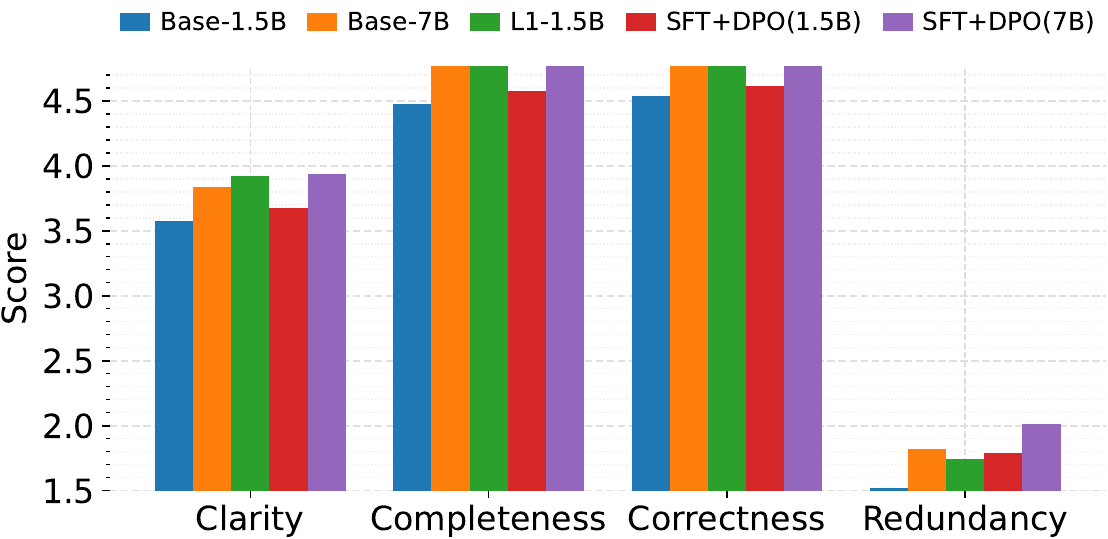}
        \caption{Unimodal}
        \label{fig:unimodal_scores}
    \end{subfigure}
    \hspace{-0.2cm}
    \begin{subfigure}[b]{0.5\textwidth}
        \includegraphics[width=\linewidth]{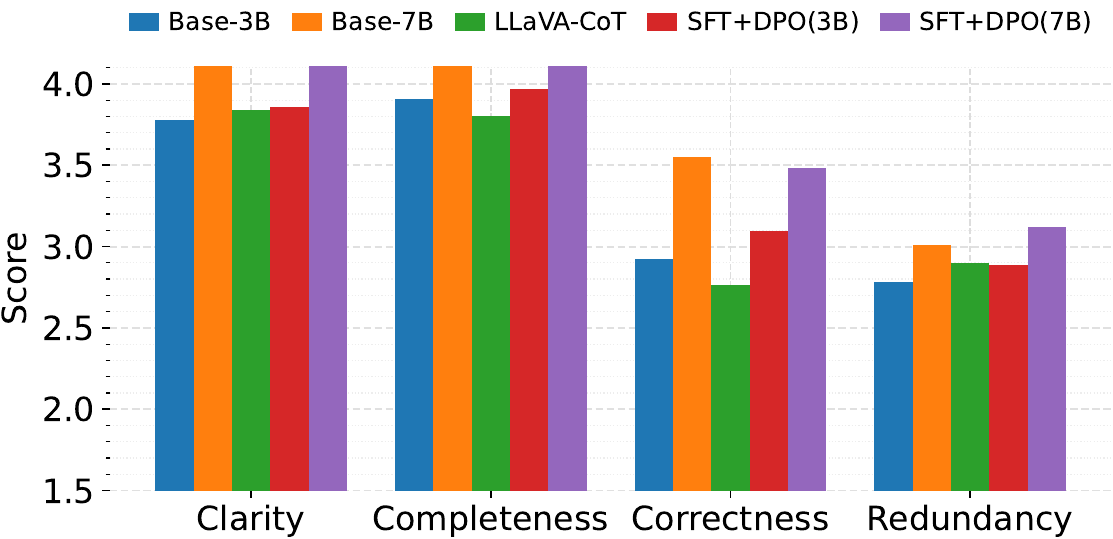}
        \caption{Multimodal}
        \label{fig:multimodal_scores}
    \end{subfigure}
   \caption{LLM-as-Judge evaluation scores for unimodal and multimodal models. Evaluations are based on a 1–5 Likert scale across four qualitative criteria—clarity, correctness, completeness, and redundancy—using GPT-4o-mini as the evaluator. SFT+DPO consistently improves both content quality and formatting over baseline models.}
    \label{fig:llm_as_judge_scores}
\end{figure*}

For unimodal models (Figure~\ref{fig:unimodal_scores}), SFT+DPO achieves the most balanced performance, scoring higher than both base and SFT-only models in correctness and completeness, while maintaining strong clarity and low redundancy. SFT alone improves clarity and reduces redundancy but lags in correctness—confirming earlier findings that it captures surface form but may sacrifice answer accuracy. DPO alone maintains correctness but shows relatively higher redundancy, suggesting limited stylistic refinement. These trends are further corroborated by the chain-level CoT scores in Figure~\ref{fig:unimodel_cotscores}, where SFT+DPO ranks highest overall.

In the multimodal setting (Figure~\ref{fig:multimodal_scores}), we observe similar patterns. The SFT+DPO models outperform others in clarity and correctness, particularly on structured reasoning tasks like MathVista and OlympiadB. SFT models again exhibit reduced verbosity, while DPO improves correctness but inherits formatting inconsistencies. As shown in Figure~\ref{fig:multimodal_cotscores}, SFT+DPO consistently yields the highest overall CoT ratings, affirming its ability to produce both accurate and readable reasoning.

Together, these results validate the effectiveness of the SFT+DPO approach in producing concise, well-structured, and accurate reasoning traces. They also highlight the importance of aligning optimization objectives with qualitative dimensions beyond final answer accuracy. We provide illustrative examples of reasoning traces generated by our models and corresponding baselines, along with commentary, in Figure~\ref{tab:unimodel_errors} and Figure~\ref{tab:multimodal_errors}, which showcase representative cases for the unimodal and multimodal setups, respectively.

% Related Work
\section{Related Work}\label{sec:related_work}

\noindent\textbf{Chain-of-Thought Reasoning.}  
Chain-of-Thought (CoT) prompting~\cite{Wei2022ChainOT} enables step-by-step reasoning in large language models (LLMs) and has been extended to zero-shot variants~\cite{kojima2022large}, self-adaptive prompting~\cite{Zhou23APO}, and complexity-based strategies~\cite{fu2023complexitybased}. Structured variants such as self-consistency~\cite{wang2022self}, Tree-of-Thought~\cite{yao2023tree}, and Graph-of-Thought~\cite{besta2023graph, Sun2024ReasoningLLM} further improve performance by enabling diverse or compositional reasoning. In multimodal settings, CoT has been adapted to enable reasoning across visual and textual inputs~\cite{zhang2023multimodal, zheng2023ddcot, pi2023visual}, enhancing performance on various tasks~\cite{liu2023visual, lu2022learn, wu2023medpalm}.

\noindent\textbf{Explicit Length-Controlled Reasoning.}  
Recent efforts aim to improve reasoning efficiency by controlling CoT length. L1~\cite{Li2024L1} uses reinforcement learning to train models for concise yet effective rationales. Other approaches use verifier-guided distillation~\cite{Zhou2024VerifierGuidedDistill} or prompt compression~\cite{Guo2024SoftCoT, wingate-etal-2022-prompt} to regulate verbosity. In multimodal domains, SCOTT~\cite{Wang2023SCOTT} and VLAA-Thinker~\cite{chen2025sftrlearlyinvestigation} leverage multi-stage summarization and derivation to reduce token usage while maintaining interpretability. These works demonstrate that reasoning length can be adapted without compromising quality.

\noindent\textbf{Efficient Thinking.}  
Efficiency in CoT generation has been explored through latent reasoning (e.g., COCONUT~\cite{Yang2024COCONUT}), mechanistic interpretability~\cite{olsson2022context, bills2023language}, and task vector methods~\cite{hendel2023task, todd2024function, huang2024multimodal}. Other strategies include activation compression~\cite{zhang2024soaring, ge2024context}, rationale distillation~\cite{Hsieh2023DistillingSbS, muennighoff2025s1}, and inference acceleration via speculative decoding~\cite{leviathan2023fast, Miao2024SEED} or parallel decoding heads~\cite{Cai2024Medusa}. Context compression techniques like Gisting~\cite{mu2024learning} and AutoCompressors~\cite{chevalier2023adapting} also improve scalability by reducing inference cost.

Our work builds on these foundations by integrating difficulty-aware CoT distillation with supervision strategies that dynamically adapt reasoning length based on task complexity. Unlike prior approaches that apply uniform compression or fixed-length control, we train models—both unimodal and multimodal—to reason proportionally to problem difficulty.

% Conclusion
\section{Conclusion}\label{sec:conclusion}

We present a scalable framework for difficulty-aware chain-of-thought (CoT) distillation that enables models to reason proportionally to problem complexity. By compressing verbose reasoning traces based on difficulty scores and combining supervised fine-tuning (SFT) with direct preference optimization (DPO), our method trains models to produce concise yet accurate reasoning. Our results demonstrate that this two-stage approach yields up to 30\% reduction in reasoning length without sacrificing final answer accuracy across both unimodal and multimodal math reasoning benchmarks. While SFT excels at learning structured and concise reasoning styles, DPO preserves correctness by aligning outputs with human preferences. Their combination captures the best of both—conciseness and correctness. Qualitative evaluations using the LLM-as-Judge framework further confirm that SFT+DPO models produce clearer, more complete, and less redundant reasoning traces. Overall, this work highlights the potential of adaptive reasoning strategies for building more efficient, interpretable, and task-aligned LLMs—models that not only solve problems but also learn how hard to think.

\section{Limitations}
While our results demonstrate the effectiveness of difficulty-aware chain-of-thought (CoT) distillation, our approach has several limitations. Our training dataset is considerably smaller than those used in prior large-scale distillation efforts such as DeepSeek-Distil and Nemotron, which may limit generalization to out-of-distribution problems. We evaluate models using the standard pass@1 metric rather than pass@k metrics that better account for generation variability. Due to compute constraints, we rely on parameter-efficient LoRA fine-tuning rather than full model fine-tuning, which may underutilize model capacity. Additionally, our difficulty estimation uses GPT-4o-mini as a proxy for true human difficulty assessment, and errors in these estimates could propagate through the distillation pipeline. These limitations highlight opportunities for further refinement and scaling of our framework in future work.

% \section{Potential Negative Impacts}

% While our research aims to improve reasoning efficiency, it carries potential risks that should be addressed. 

% Our difficulty ratings and compression approach might inadvertently encode biases about problem complexity. If certain types of math problems are systematically rated as "easier" or "harder" based on cultural or educational biases, our models may perpetuate these biases by allocating insufficient reasoning steps to problems that are actually difficult for certain user populations.

% Although our evaluation shows maintained or improved accuracy, deploying such systems in high-stakes domains (e.g., medical or financial reasoning) would require extensive testing to ensure the efficiency gains don't come at the cost of reliability in critical scenarios. While our approach demonstrates promising results on mathematical benchmarks, the balance between conciseness and thoroughness may need further calibration for domains where the costs of errors are particularly high.

% We believe these concerns can be addressed through careful deployment, continued evaluation across diverse user populations, and transparency about the limitations of difficulty-adaptive reasoning.

\newpage
\bibliographystyle{abbrvnat}
\bibliography{custom}

\begin{thebibliography}{65}
\providecommand{\natexlab}[1]{#1}
\providecommand{\url}[1]{\texttt{#1}}
\expandafter\ifx\csname urlstyle\endcsname\relax
  \providecommand{\doi}[1]{doi: #1}\else
  \providecommand{\doi}{doi: \begingroup \urlstyle{rm}\Url}\fi

\bibitem[Aggarwal and Welleck(2025)]{aggarwal2025l1controllinglongreasoning}
P.~Aggarwal and S.~Welleck.
\newblock L1: Controlling how long a reasoning model thinks with reinforcement learning, 2025.
\newblock URL \url{https://arxiv.org/abs/2503.04697}.

\bibitem[Aggarwal et~al.(2023)Aggarwal, Madaan, Yang, and Mausam]{Aggarwal2023LCPO}
P.~Aggarwal, A.~Madaan, Y.~Yang, and Mausam.
\newblock Let's sample step by step: Adaptive-consistency for efficient reasoning and coding with llms.
\newblock \emph{Proceedings of the 2023 Conference on Empirical Methods in Natural Language Processing}, pages 16624--16648, 2023.

\bibitem[Balunović et~al.(2025)Balunović, Dekoninck, Petrov, Jovanović, and Vechev]{balunovic_srimatharena_2025}
M.~Balunović, J.~Dekoninck, I.~Petrov, N.~Jovanović, and M.~Vechev.
\newblock Matharena: Evaluating llms on uncontaminated math competitions, Feb. 2025.
\newblock URL \url{https://matharena.ai/}.

\bibitem[Besta et~al.(2023)Besta, Blach, Kubicek, Gerstenberger, Gianinazzi, Gajda, Lehmann, Podstawski, Niewiadomski, Nyczyk, et~al.]{besta2023graph}
M.~Besta, N.~Blach, A.~Kubicek, R.~Gerstenberger, L.~Gianinazzi, J.~Gajda, T.~Lehmann, M.~Podstawski, H.~Niewiadomski, P.~Nyczyk, et~al.
\newblock Graph of thoughts: Solving elaborate problems with large language models.
\newblock \emph{arXiv preprint arXiv:2308.09687}, 2023.

\bibitem[Bills et~al.(2023)Bills, Jermyn, Werning, Kim, Wu, and Ellis]{bills2023language}
S.~Bills, A.~S. Jermyn, N.~Werning, J.~S. Kim, K.~Wu, and K.~Ellis.
\newblock Language models represent space and time.
\newblock \emph{Advances in Neural Information Processing Systems}, 36, 2023.

\bibitem[Boye and Moell(2025)]{boye2025largelanguagemodelsmathematical}
J.~Boye and B.~Moell.
\newblock Large language models and mathematical reasoning failures, 2025.
\newblock URL \url{https://arxiv.org/abs/2502.11574}.

\bibitem[Cai et~al.(2024)Cai, Li, Chen, Le, Yang, and Li]{Cai2024Medusa}
T.~Cai, Y.~Li, Z.~Chen, Q.~V. Le, J.~Z.~K. Yang, and C.~Li.
\newblock Medusa: Simple framework for accelerating llm generation with multiple decoding heads.
\newblock 2024.

\bibitem[Chen et~al.(2025{\natexlab{a}})Chen, Tu, Wang, Liu, Tang, Du, Zhou, and Xie]{chen2025sftrlearlyinvestigation}
H.~Chen, H.~Tu, F.~Wang, H.~Liu, X.~Tang, X.~Du, Y.~Zhou, and C.~Xie.
\newblock Sft or rl? an early investigation into training r1-like reasoning large vision-language models, 2025{\natexlab{a}}.
\newblock URL \url{https://arxiv.org/abs/2504.11468}.

\bibitem[Chen et~al.(2024)Chen, Li, Dong, Zhang, Zang, Chen, Duan, Wang, Qiao, Lin, and Zhao]{chen2024rightwayevaluatinglarge}
L.~Chen, J.~Li, X.~Dong, P.~Zhang, Y.~Zang, Z.~Chen, H.~Duan, J.~Wang, Y.~Qiao, D.~Lin, and F.~Zhao.
\newblock Are we on the right way for evaluating large vision-language models?, 2024.
\newblock URL \url{https://arxiv.org/abs/2403.20330}.

\bibitem[Chen et~al.(2025{\natexlab{b}})Chen, Xu, Liang, He, Pang, Yu, Song, Liu, Zhou, Zhang, Wang, Tu, Mi, and Yu]{chen2025think23overthinkingo1like}
X.~Chen, J.~Xu, T.~Liang, Z.~He, J.~Pang, D.~Yu, L.~Song, Q.~Liu, M.~Zhou, Z.~Zhang, R.~Wang, Z.~Tu, H.~Mi, and D.~Yu.
\newblock Do not think that much for 2+3=? on the overthinking of o1-like llms, 2025{\natexlab{b}}.
\newblock URL \url{https://arxiv.org/abs/2412.21187}.

\bibitem[Chevalier et~al.(2023)Chevalier, Wettig, Ajith, and Chen]{chevalier2023adapting}
A.~Chevalier, A.~Wettig, A.~Ajith, and D.~Chen.
\newblock Adapting language models to compress contexts.
\newblock In \emph{Proceedings of the 2023 Conference on Empirical Methods in Natural Language Processing}, pages 3829--3846, Singapore, Dec. 2023. Association for Computational Linguistics.
\newblock \doi{10.18653/v1/2023.emnlp-main.232}.
\newblock URL \url{https://aclanthology.org/2023.emnlp-main.232/}.

\bibitem[Chu et~al.(2024)Chu, Chen, Chen, Yu, He, Wang, Peng, Liu, Qin, and Liu]{chuCoTReasoningSurvey2024}
Z.~Chu, J.~Chen, Q.~Chen, W.~Yu, T.~He, H.~Wang, W.~Peng, M.~Liu, B.~Qin, and T.~Liu.
\newblock Navigate through enigmatic labyrinth a survey of chain of thought reasoning: Advances, frontiers and future.
\newblock In \emph{The 62nd Annual Meeting of the Association for Computational Linguistics: ACL 2024, Bangkok, Thailand, August 11–16, 2024}. Association for Computational Linguistics, 2024.
\newblock URL \url{https://arxiv.org/abs/2309.15402}.

\bibitem[Cobbe et~al.(2021)Cobbe, Kosaraju, Bavarian, Chen, Jun, Kaiser, Plappert, Tworek, Hilton, Nakano, Hesse, and Schulman]{cobbe2021gsm8k}
K.~Cobbe, V.~Kosaraju, M.~Bavarian, M.~Chen, H.~Jun, L.~Kaiser, M.~Plappert, J.~Tworek, J.~Hilton, R.~Nakano, C.~Hesse, and J.~Schulman.
\newblock Training verifiers to solve math word problems.
\newblock \emph{arXiv preprint arXiv:2110.14168}, 2021.

\bibitem[Damani et~al.(2024)Damani, Shenfeld, Peng, Bobu, and Andreas]{damani2024learninghardthinkinputadaptive}
M.~Damani, I.~Shenfeld, A.~Peng, A.~Bobu, and J.~Andreas.
\newblock Learning how hard to think: Input-adaptive allocation of lm computation, 2024.
\newblock URL \url{https://arxiv.org/abs/2410.04707}.

\bibitem[DeepSeek-AI(2025)]{deepseekai2025deepseekr1incentivizingreasoningcapability}
DeepSeek-AI.
\newblock Deepseek-r1: Incentivizing reasoning capability in llms via reinforcement learning, 2025.
\newblock URL \url{https://arxiv.org/abs/2501.12948}.

\bibitem[{DeepSeek-AI} et~al.(2025){DeepSeek-AI}, Guo, Yang, Zhang, Song, Zhang, Xu, Zhu, Ma, Wang, et~al.]{deepseek2025r1}
{DeepSeek-AI}, D.~Guo, D.~Yang, H.~Zhang, J.~Song, R.~Zhang, R.~Xu, Q.~Zhu, S.~Ma, P.~Wang, et~al.
\newblock Deepseek-r1: Incentivizing reasoning capability in llms via reinforcement learning.
\newblock \emph{arXiv preprint arXiv:2501.12948}, 2025.

\bibitem[Duan et~al.(2024)Duan, Yang, Qiao, Fang, Chen, Liu, Dong, Zang, Zhang, Wang, et~al.]{duan2024vlmevalkit}
H.~Duan, J.~Yang, Y.~Qiao, X.~Fang, L.~Chen, Y.~Liu, X.~Dong, Y.~Zang, P.~Zhang, J.~Wang, et~al.
\newblock Vlmevalkit: An open-source toolkit for evaluating large multi-modality models.
\newblock In \emph{Proceedings of the 32nd ACM International Conference on Multimedia}, pages 11198--11201, 2024.

\bibitem[Fu et~al.(2022)Fu, Peng, Khot, Tafjord, Clark, and Tandon]{fu2023complexitybased}
Y.~Fu, H.~Peng, T.~Khot, O.~Tafjord, P.~Clark, and N.~Tandon.
\newblock Complexity-based prompting for multi-step reasoning.
\newblock \emph{arXiv preprint arXiv:2210.00720}, 2022.

\bibitem[Ge et~al.(2024)Ge, Jing, Wang, Wang, Chen, and Wei]{ge2024context}
T.~Ge, H.~Jing, L.~Wang, X.~Wang, S.~Chen, and F.~Wei.
\newblock In-context autoencoder for context compression in a large language model.
\newblock In \emph{International Conference on Learning Representations}, 2024.

\bibitem[Gu et~al.(2025)Gu, Jiang, Shi, Tan, Zhai, Xu, Li, Shen, Ma, Liu, Wang, Zhang, Wang, Gao, Ni, and Guo]{gu2025surveyllmasajudge}
J.~Gu, X.~Jiang, Z.~Shi, H.~Tan, X.~Zhai, C.~Xu, W.~Li, Y.~Shen, S.~Ma, H.~Liu, S.~Wang, K.~Zhang, Y.~Wang, W.~Gao, L.~Ni, and J.~Guo.
\newblock A survey on llm-as-a-judge, 2025.
\newblock URL \url{https://arxiv.org/abs/2411.15594}.

\bibitem[Guo et~al.(2024)Guo, Li, Liu, Zhou, and Liao]{Guo2024SoftCoT}
J.~Guo, D.~Li, L.~Liu, P.~Zhou, and S.~Liao.
\newblock Softcot: Soft chain-of-thought for efficient reasoning with llms.
\newblock \emph{arXiv preprint arXiv:2402.10200}, 2024.

\bibitem[Hao et~al.(2023)Hao, Gu, Ma, Hong, Wang, Peng, Duan, Sun, Murphy, Lai, Wang, and Wang]{Sun2024ReasoningLLM}
S.~Hao, Y.~Gu, H.~Ma, J.~J. Hong, Z.~Wang, D.~Peng, Z.~Duan, H.~Sun, K.~P. Murphy, T.~L. Lai, A.~Wang, and L.~Wang.
\newblock Reasoning with language model is planning with world model.
\newblock \emph{arXiv preprint arXiv:2305.14992}, 2023.

\bibitem[He et~al.(2024)He, Luo, Bai, Hu, Thai, Shen, Hu, Han, Huang, Zhang, Liu, Qi, Liu, and Sun]{he2024olympiadbenchchallengingbenchmarkpromoting}
C.~He, R.~Luo, Y.~Bai, S.~Hu, Z.~L. Thai, J.~Shen, J.~Hu, X.~Han, Y.~Huang, Y.~Zhang, J.~Liu, L.~Qi, Z.~Liu, and M.~Sun.
\newblock Olympiadbench: A challenging benchmark for promoting agi with olympiad-level bilingual multimodal scientific problems, 2024.
\newblock URL \url{https://arxiv.org/abs/2402.14008}.

\bibitem[Hendel et~al.(2023)Hendel, Geva, and Globerson]{hendel2023task}
R.~Hendel, M.~Geva, and A.~Globerson.
\newblock In-context learning creates task vectors.
\newblock In \emph{Findings of the Association for Computational Linguistics: EMNLP 2023}, pages 9318--9333. Association for Computational Linguistics, 2023.

\bibitem[Hendrycks et~al.(2021)Hendrycks, Burns, Kadavath, Arora, Basart, Tang, Song, and Steinhardt]{hendrycks2021measuringmathematicalproblemsolving}
D.~Hendrycks, C.~Burns, S.~Kadavath, A.~Arora, S.~Basart, E.~Tang, D.~Song, and J.~Steinhardt.
\newblock Measuring mathematical problem solving with the math dataset, 2021.
\newblock URL \url{https://arxiv.org/abs/2103.03874}.

\bibitem[Hsieh et~al.(2023)Hsieh, Li, Yeh, Nakhost, Fujii, Ratner, Krishna, Lee, and Pfister]{Hsieh2023DistillingSbS}
C.-Y. Hsieh, C.-L. Li, C.-K. Yeh, H.~Nakhost, Y.~Fujii, A.~Ratner, R.~Krishna, C.-Y. Lee, and T.~Pfister.
\newblock Distilling step-by-step! outperforming larger language models with less training data and smaller model sizes.
\newblock \emph{arXiv preprint arXiv:2305.02301}, 2023.
\newblock URL \url{https://arxiv.org/abs/2305.02301}.

\bibitem[Huang et~al.(2024)Huang, Mitra, Arbelle, Karlinsky, Darrell, and Herzig]{huang2024multimodal}
B.~Huang, C.~Mitra, A.~Arbelle, L.~Karlinsky, T.~Darrell, and R.~Herzig.
\newblock Multimodal task vectors enable many-shot multimodal in-context learning.
\newblock \emph{arXiv preprint arXiv:2406.15334}, 2024.

\bibitem[Kojima et~al.(2022)Kojima, Gu, Reid, Matsuo, and Iwasawa]{kojima2022large}
T.~Kojima, S.~S. Gu, M.~Reid, Y.~Matsuo, and Y.~Iwasawa.
\newblock Large language models are zero-shot reasoners.
\newblock \emph{Advances in Neural Information Processing Systems}, 35:\penalty0 22199--22213, 2022.

\bibitem[Lang et~al.(2024)Lang, Ji, Wang, Yao, Xu, Yan, Gui, and Zhang]{Yang2024COCONUT}
Y.~Lang, Z.~Ji, H.~Wang, W.~Yao, J.~Xu, M.~Yan, T.~Gui, and Q.~Zhang.
\newblock Coconut: Combining implicit and explicit knowledge for chain-of-thought reasoning.
\newblock 2024.

\bibitem[Leviathan et~al.(2023)Leviathan, Kalman, and Matias]{leviathan2023fast}
Y.~Leviathan, M.~Kalman, and Y.~Matias.
\newblock Fast inference from transformers via speculative decoding.
\newblock 2023.

\bibitem[Li et~al.(2024)Li, Chen, Kolter, Parameswaran, Neubig, and Jiang]{Li2024L1}
H.~Li, Z.~Chen, J.~Z. Kolter, A.~Parameswaran, G.~Neubig, and Y.~A. Jiang.
\newblock L1: Controlling how long a reasoning model thinks with reinforcement learning.
\newblock \emph{arXiv preprint arXiv:2503.04697}, 2024.

\bibitem[Liu et~al.(2023)Liu, Chen, Lin, Tegmark, and Zhou]{liu2023visual}
Z.~Liu, Y.~Chen, Y.~Lin, M.~Tegmark, and B.~Zhou.
\newblock Visual chain of thought: Bridging logical gaps with multimodal infillings.
\newblock \emph{arXiv preprint arXiv:2305.09666}, 2023.

\bibitem[Lu et~al.(2022)Lu, Mishra, Xia, Qiu, Chang, Zhu, Tafjord, Clark, and Kalyan]{lu2022learn}
P.~Lu, S.~Mishra, T.~Xia, L.~Qiu, K.-W. Chang, S.-C. Zhu, O.~Tafjord, P.~Clark, and A.~Kalyan.
\newblock Learn to explain: Multimodal reasoning via thought chains for science question answering.
\newblock \emph{Advances in Neural Information Processing Systems}, 35:\penalty0 2507--2522, 2022.

\bibitem[Miao et~al.(2024)Miao, Ye, Wang, and Huang]{Miao2024SEED}
C.~Miao, M.~Ye, F.~Wang, and F.~Huang.
\newblock Seed: Accelerating reasoning tree construction via scheduled speculative decoding.
\newblock 2024.

\bibitem[Moshkov et~al.(2025)Moshkov, Hanley, Sorokin, Toshniwal, Henkel, Schifferer, Du, and Gitman]{moshkov2025aimo2}
I.~Moshkov, D.~Hanley, I.~Sorokin, S.~Toshniwal, C.~Henkel, B.~Schifferer, W.~Du, and I.~Gitman.
\newblock Aimo-2 winning solution: Building state-of-the-art mathematical reasoning models with openmathreasoning dataset.
\newblock \emph{arXiv preprint arXiv:2504.16891}, 2025.

\bibitem[Mu et~al.(2024)Mu, Li, and Goodman]{mu2024learning}
J.~Mu, X.~L. Li, and N.~D. Goodman.
\newblock Learning to compress prompts with gist tokens.
\newblock \emph{Advances in Neural Information Processing Systems}, 36, 2024.

\bibitem[Muennighoff et~al.(2025)Muennighoff, Yang, Shi, Li, Fei-Fei, Hajishirzi, Zettlemoyer, Liang, Cand{\`e}s, and Hashimoto]{muennighoff2025s1}
N.~Muennighoff, Z.~Yang, W.~Shi, X.~L. Li, L.~Fei-Fei, H.~Hajishirzi, L.~Zettlemoyer, P.~Liang, E.~Cand{\`e}s, and T.~Hashimoto.
\newblock s1: Simple test-time scaling.
\newblock \emph{arXiv preprint arXiv:2501.19393}, 2025.

\bibitem[Olsson et~al.(2022)Olsson, Elhage, Nanda, Joseph, Hao, Henighan, DasSarma, Bai, Hernandez, Askell, et~al.]{olsson2022context}
C.~Olsson, N.~Elhage, N.~Nanda, N.~Joseph, N.~Hao, T.~Henighan, A.~DasSarma, Y.~Bai, D.~Hernandez, A.~Askell, et~al.
\newblock In-context learning and induction heads.
\newblock In \emph{The Tenth International Conference on Learning Representations (ICLR)}, 2022.

\bibitem[OpenAI(2024)]{openai2024openaio1card}
OpenAI.
\newblock Openai o1 system card, 2024.
\newblock URL \url{https://arxiv.org/abs/2412.16720}.

\bibitem[Patil(2025)]{patil2025advancingreasoninglargelanguage}
A.~Patil.
\newblock Advancing reasoning in large language models: Promising methods and approaches, 2025.
\newblock URL \url{https://arxiv.org/abs/2502.03671}.

\bibitem[Phan et~al.(2025)Phan, Gatti, and Others]{phan2025humanitysexam}
L.~Phan, A.~Gatti, and Others.
\newblock Humanity's last exam, 2025.
\newblock URL \url{https://arxiv.org/abs/2501.14249}.

\bibitem[Pi et~al.(2023)Pi, Wang, Zhang, Fu, Fu, Lo, and Zou]{pi2023visual}
X.~Pi, F.~Wang, S.~Zhang, C.~Fu, C.~Fu, E.~Lo, and Y.~Zou.
\newblock Visual chain-of-thought: Bridging logical gaps with multimodal infillings.
\newblock \emph{arXiv preprint arXiv:2305.02317}, 2023.

\bibitem[Rafailov et~al.(2024)Rafailov, Sharma, Mitchell, Ermon, Manning, and Finn]{rafailov2024directpreferenceoptimizationlanguage}
R.~Rafailov, A.~Sharma, E.~Mitchell, S.~Ermon, C.~D. Manning, and C.~Finn.
\newblock Direct preference optimization: Your language model is secretly a reward model, 2024.
\newblock URL \url{https://arxiv.org/abs/2305.18290}.

\bibitem[Reid et~al.(2023)Reid, Zhou, Fu, Liu, and Neubig]{Zhou23APO}
M.~Reid, D.~Zhou, Y.~Fu, P.~Liu, and G.~Neubig.
\newblock Automatic prompt optimization with "gradient descent" and beam search.
\newblock In \emph{Findings of the Association for Computational Linguistics: EMNLP 2023}, 2023.
\newblock URL \url{https://aclanthology.org/2023.findings-emnlp.1001}.

\bibitem[Samsi et~al.(2023)Samsi, Zhao, McDonald, Li, Michaleas, Jones, Bergeron, Kepner, Tiwari, and Gadepally]{samsi2023words}
S.~Samsi, D.~Zhao, J.~McDonald, B.~Li, A.~Michaleas, M.~Jones, W.~Bergeron, J.~Kepner, D.~Tiwari, and V.~Gadepally.
\newblock From words to watts: Benchmarking the energy costs of large language model inference.
\newblock In \emph{2023 IEEE High Performance Extreme Computing Conference (HPEC)}, pages 1--9. IEEE, 2023.

\bibitem[Team(2025{\natexlab{a}})]{sky_t1_2025}
N.~Team.
\newblock Sky-t1: Train your own o1 preview model within \$450.
\newblock https://novasky-ai.github.io/posts/sky-t1, 2025{\natexlab{a}}.
\newblock Accessed: 2025-01-09.

\bibitem[Team(2025{\natexlab{b}})]{qwq32b}
Q.~Team.
\newblock Qwq-32b: Embracing the power of reinforcement learning, March 2025{\natexlab{b}}.
\newblock URL \url{https://qwenlm.github.io/blog/qwq-32b/}.

\bibitem[Todd et~al.(2024)Todd, Li, Sharma, Mueller, Wallace, and Bau]{todd2024function}
E.~Todd, M.~L. Li, A.~S. Sharma, A.~Mueller, B.~C. Wallace, and D.~Bau.
\newblock Function vectors in large language models.
\newblock In \emph{Proceedings of the 2024 International Conference on Learning Representations}, 2024.

\bibitem[Wang et~al.(2024)Wang, Pan, Shi, Lu, Ren, Zhou, Zhan, and Li]{wang2024measuring}
K.~Wang, J.~Pan, W.~Shi, Z.~Lu, H.~Ren, A.~Zhou, M.~Zhan, and H.~Li.
\newblock Measuring multimodal mathematical reasoning with math-vision dataset.
\newblock In \emph{The Thirty-eight Conference on Neural Information Processing Systems Datasets and Benchmarks Track}, 2024.
\newblock URL \url{https://openreview.net/forum?id=QWTCcxMpPA}.

\bibitem[Wang et~al.(2023)Wang, Wang, Li, Gao, Yin, and Wei]{Wang2023SCOTT}
P.~Wang, Z.~Wang, Z.~Li, Y.~Gao, B.~Yin, and X.~Wei.
\newblock Scott: Self-consistent chain-of-thought distillation.
\newblock 2023.

\bibitem[Wang et~al.(2022)Wang, Wei, Schuurmans, Le, Chi, and Zhou]{wang2022self}
X.~Wang, J.~Wei, D.~Schuurmans, Q.~Le, E.~Chi, and D.~Zhou.
\newblock Self-consistency improves chain of thought reasoning in language models.
\newblock \emph{arXiv preprint arXiv:2203.11171}, 2022.

\bibitem[Wei et~al.(2022)Wei, Wang, Schuurmans, Bosma, Ichter, Xia, Chi, Le, and Zhou]{Wei2022ChainOT}
J.~Wei, X.~Wang, D.~Schuurmans, M.~Bosma, B.~Ichter, F.~Xia, E.~Chi, Q.~Le, and D.~Zhou.
\newblock Chain-of-thought prompting elicits reasoning in large language models.
\newblock \emph{Advances in Neural Information Processing Systems}, 35:\penalty0 24824--24837, 2022.

\bibitem[Wingate et~al.(2022)Wingate, Shoeybi, and Sorensen]{wingate-etal-2022-prompt}
D.~Wingate, M.~Shoeybi, and T.~Sorensen.
\newblock Prompt compression and contrastive conditioning for controllability and toxicity reduction in language models.
\newblock In \emph{Findings of the Association for Computational Linguistics: EMNLP 2022}, pages 5621--5634, Abu Dhabi, United Arab Emirates, Dec. 2022. Association for Computational Linguistics.
\newblock \doi{10.18653/v1/2022.findings-emnlp.412}.
\newblock URL \url{https://aclanthology.org/2022.findings-emnlp.412/}.

\bibitem[Wolf et~al.(2020)Wolf, Debut, Sanh, Chaumond, Delangue, Moi, Cistac, Rault, Louf, Funtowicz, Davison, Shleifer, von Platen, Ma, Jernite, Plu, Xu, Scao, Gugger, Drame, Lhoest, and Rush]{wolf-etal-2020-transformers}
T.~Wolf, L.~Debut, V.~Sanh, J.~Chaumond, C.~Delangue, A.~Moi, P.~Cistac, T.~Rault, R.~Louf, M.~Funtowicz, J.~Davison, S.~Shleifer, P.~von Platen, C.~Ma, Y.~Jernite, J.~Plu, C.~Xu, T.~L. Scao, S.~Gugger, M.~Drame, Q.~Lhoest, and A.~M. Rush.
\newblock Transformers: State-of-the-art natural language processing.
\newblock In \emph{Proceedings of the 2020 Conference on Empirical Methods in Natural Language Processing: System Demonstrations}, pages 38--45, Online, Oct. 2020. Association for Computational Linguistics.
\newblock URL \url{https://www.aclweb.org/anthology/2020.emnlp-demos.6}.

\bibitem[Wu et~al.(2023)Wu, Cao, He, Pudipeddi, Ghanathe, Tantia, Thaker, Liu, Nagrani, Kejriwal, et~al.]{wu2023medpalm}
K.~Wu, J.~J. Cao, T.~He, R.~Pudipeddi, P.~Ghanathe, P.~Tantia, V.~Thaker, Y.~Liu, A.~Nagrani, S.~Kejriwal, et~al.
\newblock Med-palm 2: Towards expert-level medical question answering with large language models.
\newblock \emph{arXiv preprint arXiv:2305.09617}, 2023.

\bibitem[Xu et~al.(2024)Xu, Jin, Li, Song, Sun, and Yuan]{xu2024llavacotletvisionlanguage}
G.~Xu, P.~Jin, H.~Li, Y.~Song, L.~Sun, and L.~Yuan.
\newblock Llava-cot: Let vision language models reason step-by-step, 2024.
\newblock URL \url{https://arxiv.org/abs/2411.10440}.

\bibitem[Yang et~al.(2024)Yang, Yang, Zhang, Hui, Zheng, Yu, Li, Liu, Huang, Wei, Lin, Yang, Tu, Zhang, Yang, Yang, Zhou, Lin, Dang, Lu, Bao, Yang, Yu, Li, Xue, Zhang, Zhu, Men, Lin, Li, Tang, Xia, Ren, Ren, Fan, Su, Zhang, Wan, Liu, Cui, Zhang, and Qiu]{qwen2.5}
A.~Yang, B.~Yang, B.~Zhang, B.~Hui, B.~Zheng, B.~Yu, C.~Li, D.~Liu, F.~Huang, H.~Wei, H.~Lin, J.~Yang, J.~Tu, J.~Zhang, J.~Yang, J.~Yang, J.~Zhou, J.~Lin, K.~Dang, K.~Lu, K.~Bao, K.~Yang, L.~Yu, M.~Li, M.~Xue, P.~Zhang, Q.~Zhu, R.~Men, R.~Lin, T.~Li, T.~Tang, T.~Xia, X.~Ren, X.~Ren, Y.~Fan, Y.~Su, Y.~Zhang, Y.~Wan, Y.~Liu, Z.~Cui, Z.~Zhang, and Z.~Qiu.
\newblock Qwen2.5 technical report.
\newblock \emph{arXiv preprint arXiv:2412.15115}, 2024.

\bibitem[Yao et~al.(2023)Yao, Yu, Zhao, Shafran, Griffith, Xu, and Shen]{yao2023tree}
S.~Yao, D.~Yu, J.~Zhao, I.~Shafran, T.~L. Griffith, Y.~Xu, and X.~Shen.
\newblock Tree of thoughts: Deliberate problem solving with large language models.
\newblock \emph{arXiv preprint arXiv:2305.10601}, 2023.

\bibitem[Zhang et~al.(2024{\natexlab{a}})Zhang, Liu, Xiao, Shao, Ye, and Dou]{zhang2024soaring}
P.~Zhang, Z.~Liu, S.~Xiao, N.~Shao, Q.~Ye, and Z.~Dou.
\newblock Soaring from 4k to 400k: Extending llm's context with activation beacon.
\newblock \emph{arXiv preprint arXiv:2401.03462}, 2024{\natexlab{a}}.

\bibitem[Zhang et~al.(2024{\natexlab{b}})Zhang, Jiang, Zhang, Lin, Guo, Qiu, Zhou, Lu, Chang, Gao, and Li]{zhang2024mathversedoesmultimodalllm}
R.~Zhang, D.~Jiang, Y.~Zhang, H.~Lin, Z.~Guo, P.~Qiu, A.~Zhou, P.~Lu, K.-W. Chang, P.~Gao, and H.~Li.
\newblock Mathverse: Does your multi-modal llm truly see the diagrams in visual math problems?, 2024{\natexlab{b}}.
\newblock URL \url{https://arxiv.org/abs/2403.14624}.

\bibitem[Zhang et~al.(2023)Zhang, Zhang, Li, Zhao, Karypis, and Smola]{zhang2023multimodal}
Z.~Zhang, A.~Zhang, M.~Li, H.~Zhao, G.~Karypis, and A.~Smola.
\newblock Multimodal chain-of-thought reasoning in language models.
\newblock \emph{arXiv preprint arXiv:2302.00923}, 2023.

\bibitem[Zheng et~al.(2023{\natexlab{a}})Zheng, Yang, Tang, Zhou, and Yang]{zheng2023ddcot}
G.~Zheng, B.~Yang, J.~Tang, H.-Y. Zhou, and S.~Yang.
\newblock Ddcot: Duty-distinct chain-of-thought prompting for multimodal reasoning in language models.
\newblock \emph{arXiv preprint arXiv:2310.16436}, 2023{\natexlab{a}}.

\bibitem[Zheng et~al.(2023{\natexlab{b}})Zheng, Chiang, Sheng, Zhuang, Wu, Zhuang, Lin, Li, Li, Xing, Zhang, Gonzalez, and Stoica]{zheng2023judgingllmasajudgemtbenchchatbot}
L.~Zheng, W.-L. Chiang, Y.~Sheng, S.~Zhuang, Z.~Wu, Y.~Zhuang, Z.~Lin, Z.~Li, D.~Li, E.~P. Xing, H.~Zhang, J.~E. Gonzalez, and I.~Stoica.
\newblock Judging llm-as-a-judge with mt-bench and chatbot arena, 2023{\natexlab{b}}.
\newblock URL \url{https://arxiv.org/abs/2306.05685}.

\bibitem[Zheng et~al.(2024)Zheng, Zhang, Zhang, Ye, Luo, Feng, and Ma]{zheng2024llamafactory}
Y.~Zheng, R.~Zhang, J.~Zhang, Y.~Ye, Z.~Luo, Z.~Feng, and Y.~Ma.
\newblock Llamafactory: Unified efficient fine-tuning of 100+ language models.
\newblock In \emph{Proceedings of the 62nd Annual Meeting of the Association for Computational Linguistics (Volume 3: System Demonstrations)}, Bangkok, Thailand, 2024. Association for Computational Linguistics.
\newblock URL \url{http://arxiv.org/abs/2403.13372}.

\bibitem[Zhou et~al.(2024)Zhou, Lin, Yasunaga, and Ren]{Zhou2024VerifierGuidedDistill}
P.~Y. Zhou, B.~Y. Lin, M.~Yasunaga, and X.~Ren.
\newblock Distilling multi-step reasoning capabilities of large language models into smaller models via verifier-guided sequence generation.
\newblock 2024.

\end{thebibliography}

%%%%%%%%%%%%%%%%%%%%%%%%%%%%%%%%%%%%%%%%%%%%%%%%%%%%%%%%%%%%
\appendix
\section{Appendix}

\subsection{Hyperparameters}\label{appsubsec:hyperparams}
In Table~\ref{tab:hyperparams} we provide a detailed list of parameters used in our experiments.
\clearpage
\begin{table*}[p]
\centering
\label{tab:hyperparameters}
\vspace{0.5cm}
\begin{tabular}{p{5cm}>{\centering\arraybackslash}p{4cm}>{\centering\arraybackslash}p{4cm}}
\toprule
\textbf{Training Phase} & \multicolumn{1}{c}{\textbf{R1-Distill-Qwen-1.5B}} & \multicolumn{1}{c}{\textbf{R1-Distill-Qwen-7B}} \\
 & \multicolumn{1}{c}{\textbf{Qwen2.5-VL-3B}} & \multicolumn{1}{c}{\textbf{Qwen2.5-VL-7B-Instruct}} \\
\midrule
\multicolumn{3}{l}{\textbf{Supervised Fine-Tuning (SFT)}} \\
\midrule
LoRA rank & 16 & 16 \\
LoRA target & all & all \\
Freeze vision tower & True & True \\
Freeze multi-modal projector & True & True \\
Cutoff length & 4096 & 4096 \\
Per device batch size & 2 & 2 \\
Gradient accumulation steps & 8 & 8 \\
Effective batch size & 128 & 128 \\
Learning rate & 3.0e-4 & 2.0e-4 \\
Training epochs & 3.0 & 3.0 \\
\midrule
\multicolumn{3}{l}{\textbf{Direct Preference Optimization (DPO)}} \\
\midrule
LoRA rank & 16 & 16 \\
LoRA target & all & all \\
Pref beta & 0.1 & 0.1 \\
Pref loss & sigmoid & sigmoid \\
Freeze vision tower & True & True \\
Freeze multi-modal projector & True & True \\
Train MM proj only & False & False \\
Cutoff length & 4096 & 4096 \\
Per device batch size & 1 & 1 \\
Gradient accumulation steps & 8 & 8 \\
Effective batch size & 64 & 64 \\
Learning rate & 7.0e-6 & 5.0e-6 \\
Training epochs & 3.0 & 3.0 \\
LR scheduler type & cosine & cosine \\
Warmup ratio & 0.05 & 0.05 \\
\bottomrule
\end{tabular}
\caption{Hyperparameters used for model training}
\label{tab:hyperparams}
\end{table*}
\clearpage

\subsection{Prompts}\label{appsubsec:prompts}
In this section, we present the complete set of prompts used across the various tasks in our study.
\subsubsection{Rating}\label{appsubsubsec:rating}
The prompts used to estimate difficulty are shown in Figure~\ref{fig:unimodal_difficulty_prompt} and Figure~\ref{fig:multimodal_difficulty_prompt}, corresponding to the unimodal and multimodal setups, respectively.
\begin{figure*}[h]
    \centering
    \tcbset{colframe=black, colback=gray!10, arc=5mm}
    \begin{tcolorbox}
    \small
    \textbf{You will be given a math problem. Your job is to grade the difficulty level from 1-10 according to the AoPS standard.} \\

    \textbf{Here is the standard:} \\
    All levels are estimated and refer to averages. The following is a rough standard based on the USA tier system AMC 8 - AMC 10 - AMC 12 - AIME - USAMO/USAJMO - IMO, representing Middle School - Junior High - High School - Challenging High School - Olympiad levels. Other contests can be interpolated against this. \\

    \textbf{Notes:} \\
    - Multiple choice tests like AMC are rated as though they are free-response. \\
    - Some Olympiads are taken in 2 sessions, with similarly difficult sets of questions. \\

    \textbf{Scale} \\
    1: Problems strictly for beginners (e.g., AMC 8 1-20, AMC 10 1-10). \\
    2: For motivated beginners (e.g., AMC 8 21-25, AMC 10 11-20). \\
    3: Advanced beginner problems (e.g., AMC 10 21-25, AIME 4-6). \\
    4: Intermediate-level problems (e.g., AMC 12 21-25, AIME 7-9). \\
    5: More difficult AIME problems (10-12), simple Olympiad problems. \\
    6: High-level AIME (13-15), introductory Olympiad questions. \\
    7: Tougher Olympiad questions requiring technical knowledge. \\
    8: High-level Olympiad questions. \\
    9: Expert Olympiad questions. \\
    10: Extremely difficult problems unsuitable for standard competitions. \\

    \textbf{Examples} \\
    <1: Counting edges/corners/faces of a cube (2003 AMC 8 Problem 1)>. \\
    1: How many integers satisfy $|x| < 3\pi$? (2021 AMC 10B Problem 1). \\
    2: Probability problem with die rolls (2021 AMC 10B Problem 18). \\
    3: Geometry problem with area calculation (2018 AMC 10A Problem 24). \\
    4: Recursive sequence problem (2019 AMC 10B Problem 24). \\
    5: System of equations (JBMO 2020/1). \\
    6: Complex geometry problem (2020 AIME I Problem 15). \\
    7: IMO 2015 Problem 1 about balanced sets. \\
    8: IMO 2014 Problem 5 about coin denominations. \\
    9: IMO 2022 Problem 3 about prime arrangements. \\
    10: IMO 2020 Problem 6 about point separation. \\

    \textbf{Task:} \\
    Problem to be labeled: \{problem\}. \\
    Please put your estimation of the difficulty inside [[level]]. \\

    \textbf{Important:} You should place the difficulty from 1-10 into the [[]], not the solution of the problem. \\

    \textbf{Output:} [[insert your difficulty level here]]
    \end{tcolorbox}
    \caption{Math problem difficulty grading prompt. The task is to classify math problems into difficulty levels 1-10 according to AoPS standards, with examples provided for each level.}
    \label{fig:unimodal_difficulty_prompt}
\end{figure*}

\begin{figure*}[h]
    \centering
    \tcbset{colframe=black, colback=gray!10, arc=5mm}
    \begin{tcolorbox}
    \small
    \textbf{You will be given a multimodal math problem, which includes mathematical expressions and/or visual elements (such as graphs, diagrams, or charts). Your task is to grade the difficulty level from 1-10 according to the AoPS standard.} \\

    \textbf{Here is the standard:} \\
    All levels are estimated and refer to averages. The following is a rough standard based on the USA tier system AMC 8 - AMC 10 - AMC 12 - AIME - USAMO/USAJMO - IMO, representing Middle School - Junior High - High School - Challenging High School - Olympiad levels. Other contests can be interpolated against this. \\

    \textbf{Notes:} \\
    - Multiple choice tests like AMC are rated as though they are free-response. Test-takers can use answer choices as hints. \\
    - Some Olympiads are taken in 2 sessions, with similarly difficult sets of questions numbered as one sequence. \\

    \textbf{Scale} \\
    1: Strictly beginner (MOEMS, AMC 8 1-20, standard middle school problems). \\
    2: Motivated beginners (AMC 8 21-25, AMC 10 11-20, complex word problems). \\
    3: Creative thinking required (AMC 10 21-25, AIME 4-6). \\
    4: Intermediate (AMC 12 21-25, AIME 7-9). \\
    5: Difficult AIME (10-12), simple Olympiad proofs. \\
    6: High AIME (13-15), introductory Olympiad. \\
    7: Technical Olympiad questions (USAJMO 3/6, medium IMO 2/5). \\
    8: Advanced Olympiad (hard IMO 2/5, easiest IMO 3/6). \\
    9: Expert Olympiad (average IMO 3/6). \\
    10: Extremely tedious/difficult (beyond standard competitions). \\

    \textbf{Examples} \\
    <1: Counting edges/corners/faces of a cube (2003 AMC 8 Problem 1)>. \\
    1: Integer solutions to $|x| < 3\pi$ (2021 AMC 10B Problem 1). \\
    2: Die roll probability with even number conditions (2021 AMC 10B Problem 18). \\
    3: Triangle area with midpoints and angle bisectors (2018 AMC 10A Problem 24). \\
    4: Recursive sequence interval analysis (2019 AMC 10B Problem 24). \\
    5: System with reciprocal equations (JBMO 2020/1). \\
    6: Acute triangle geometry with circumcircles (2020 AIME I Problem 15). \\
    7: Balanced set existence proof (IMO 2015 Problem 1). \\
    8: Coin collection partitioning proof (IMO 2014 Problem 5). \\
    9: Prime circle arrangement uniqueness (IMO 2022 Problem 3). \\
    10: Point separation line existence proof (IMO 2020 Problem 6). \\

    \textbf{Task:} \\
    Problem to be labeled: \{problem\}. \\
    Consider both mathematical content and visual elements. Place difficulty in [[level]]. \\

    \textbf{Important:} \\
    - Only output the numerical rating in [[ ]]. \\
    - Account for answer choice hints in multiple-choice problems. \\
    
    \textbf{Output:} [[insert your difficulty level here]]
    \end{tcolorbox}
    \caption{Multimodal math problem difficulty grading prompt. Extends AoPS standards to problems combining mathematical expressions with visual elements, with explicit guidance for handling multiple-choice hints and graphical components.}
    \label{fig:multimodal_difficulty_prompt}
\end{figure*}

\subsubsection{Chain-of-Thought Compression}\label{appsubsubsec:compression}
The prompts used to compress long chain-of-thought reasoning are shown in Figure~\ref{fig:unimodal_compression_prompt} and Figure~\ref{fig:multimodal_compression_prompt}, corresponding to the unimodal and multimodal setups, respectively.
\begin{figure*}[h]
    \centering
    \tcbset{colframe=black, colback=gray!10, arc=5mm}
    \begin{tcolorbox}
    \small
    \textbf{You are given a detailed math problem solution, where the full chain-of-thought (CoT) reasoning is enclosed within <think> and </think> tags. Your task is to refine this reasoning to improve clarity, conciseness, and logical flow, while preserving all essential mathematical steps and the original problem-solving structure.}

    \textbf{Your objective is to:}
    - Eliminate redundancy, verbosity, and non-essential commentary  
    - Remove unnecessary explanations or filler that do not contribute to solving the problem  
    - Preserve all mathematically necessary steps, notation, and logical transitions  
    - Streamline the reasoning while respecting the original depth and complexity  

    \textbf{Instructions:}
    The refinement should be proportional to the problem’s difficulty (a higher difficulty rating indicates a harder problem). If the original solution is long and overly detailed, focus on removing only redundant or verbose elements. Do not oversimplify or compress steps aggressively — ensure the solution remains complete, accurate, and easy to follow. The difficulty rating is provided on a scale from 1 to 10, where 1 is the easiest and 10 is the hardest.

    \textbf{Task Guidelines:}
    - Retain all mathematically necessary steps, logical transitions, and essential details required to understand and solve the problem correctly
    - Only restructure or rephrase if it clearly improves clarity without altering the logic or meaning
    - Eliminate redundancy, verbose phrasing, filler commentary, or repeated logic that does not contribute directly to problem-solving
    - Maintain mathematical accuracy, coherence, and the original formatting style
    - The degree of refinement (i.e., the length and detail of the refined CoT) should be proportional to the problem’s difficulty — simpler problems should be more concise, while complex ones may require more detailed reasoning to arrive at the solution

    \textbf{Output Format:}
    1. Enclose the refined reasoning inside <think> and </think> tags  
    2. On a \textbf{new line} after </think>, write the final answer using \boxed{{final\_answer}}

    \textbf{Input:}  
    Problem Difficulty: \{rating\}  
    Solution:  
    \{solution\}

    \textbf{Output:}  
    <think>  
    \{\{refined CoT reasoning\}\}  
    </think>  
    \boxed{\{final\_answer\}}
    \end{tcolorbox}
    \caption{Chain-of-thought refinement prompt for mathematical solutions. Specifies verbosity reduction rules while preserving essential logic and difficulty-proportional editing.}
    \label{fig:unimodal_compression_prompt}
\end{figure*}
\begin{figure*}[h]
    \centering
    \tcbset{colframe=black, colback=gray!10, arc=5mm}
    \begin{tcolorbox}
    \small
    \textbf{You are given a detailed multimodal math problem solution, where the full chain-of-thought (CoT) reasoning is provided. Your task is to refine this reasoning to improve clarity, conciseness, and logical flow, while preserving all essential mathematical steps and the original problem-solving structure.}

    \textbf{Your objective is to:}
    - Eliminate redundancy, verbosity, and non-essential commentary  
    - Remove unnecessary explanations or filler that do not contribute to solving the problem  
    - Preserve all mathematically necessary steps, notation, and logical transitions  
    - Streamline the reasoning while respecting the original depth and complexity  

    \textbf{Instructions:}
    The refinement should be proportional to the problem’s difficulty (a higher difficulty rating indicates a harder problem). If the original solution is long and overly detailed, focus on removing only redundant or verbose elements. Do not oversimplify or compress steps aggressively — ensure the solution remains complete, accurate, and easy to follow. The difficulty rating is provided on a scale from 1 to 10, where 1 is the easiest and 10 is the hardest.

    \textbf{Task Guidelines:}
    - Retain all mathematically necessary steps, logical transitions, and essential details required to understand and solve the problem correctly
    - Only restructure or rephrase if it clearly improves clarity without altering the logic or meaning
    - Eliminate redundancy, verbose phrasing, filler commentary, or repeated logic that does not contribute directly to problem-solving
    - Maintain mathematical accuracy, coherence, and the original formatting style
    - The degree of refinement (i.e., the length and detail of the refined CoT) should be proportional to the problem’s difficulty — simpler problems should be more concise, while complex ones may require more detailed reasoning to arrive at the solution

    \textbf{Output Format:}
    1. Enclose the refined reasoning inside <think> and </think> tags  
    2. On a \textbf{new line} after </think>, write the final answer using \boxed{{final\_answer}}

    \textbf{Input:}  
    Problem Difficulty: \{rating\}  
    Solution:  
    \{solution\}

    \textbf{Output:}  
    <think>  
    \{\{refined CoT reasoning\}\}  
    </think>  
    \boxed{\{final\_answer\}}
    \end{tcolorbox}
    \caption{Multimodal chain-of-thought refinement prompt. Maintains identical structure to mathematical version while implicitly handling visual elements through preserved problem-solving structure.}
    \label{fig:multimodal_compression_prompt}
\end{figure*}

\subsubsection{Verification}\label{appsubsubsec:verification}
The prompt used to verify the multimodal answers is shown in Figure~\ref{fig:verification_prompt}.
\begin{figure*}[h]
    \centering
    \tcbset{colframe=black, colback=gray!10, arc=5mm}
    \begin{tcolorbox}
    \small
    \textbf{You are given a question, the correct answer, and a candidate solution. Your task is to verify if the candidate solution is correct by comparing the final answer in the candidate solution with the correct answer.}

    Question: \{question\}

    Correct Answer: \{answer\}

    Solution: \{solution\}

    \textbf{Instruction:} Verify whether the final answer in the candidate solution matches the correct answer. Provide your response in the format \(\boxed{\text{correct}}\) if the candidate solution is correct, or \(\boxed{\text{incorrect}}\) if it is incorrect. *Do not provide any additional information*.

    \textbf{Final Answer:} \(\boxed{\text{correct}}\) or \(\boxed{\text{incorrect}}\)
    \end{tcolorbox}
    \caption{Answer verification prompt. Directly replicates original instructions for strict answer matching without modifications.}
    \label{fig:verification_prompt}
\end{figure*}

\subsubsection{Multimodal Evaluation}\label{appsubsubsec:multimodal_evaluation}
Figure~\ref{fig:multimodal_eval_base_prompt} shows the base prompt used to evaluate all multimodal models. The prompts for the zero-shot direct answer, zero-shot chain-of-thought, and long chain-of-thought setups are shown in Figures~\ref{fig:multimodal_eval_direct_answer_prompt},~\ref{fig:multimodal_eval_zero_shot_cot_prompt}, and~\ref{fig:multimodal_eval_long_cot_prompt}, respectively.
\begin{figure*}[h]
    \centering
    \tcbset{colframe=black, colback=gray!10, arc=5mm}
    \begin{tcolorbox}
    \small
    \textbf{Please answer the question below, explaining your reasoning step by step before providing the final answer.}

    Question:
    \{question\}
    \end{tcolorbox}
    \caption{Step-by-step explanation prompt. Directly replicates original instructions for detailed reasoning before final answer.}
    \label{fig:multimodal_eval_base_prompt}
\end{figure*}

\begin{figure*}[h]
    \centering
    \tcbset{colframe=black, colback=gray!10, arc=5mm}
    \begin{tcolorbox}
    \small
    \textbf{You are given a math question and your task is to provide the final answer in the format \(\boxed{\{answer\}}\). *Do not include any explanations, reasoning, intermediate steps, or additional information in your response.*}

    Question: <image> \{question\}

    Answer:
    \end{tcolorbox}
    \caption{Answer submission prompt. Directly replicates original instructions for answer-only responses with strict formatting requirements.}
    \label{fig:multimodal_eval_direct_answer_prompt}
\end{figure*}
\begin{figure*}[h]
    \centering
    \tcbset{colframe=black, colback=gray!10, arc=5mm}
    \begin{tcolorbox}
    \small
    \textbf{You are given a math question, and your task is to solve it. *Think step by step* before providing your final answer in the format \(\boxed{\{answer\}}\).}

    Question: <image> \{question\}

    Solution:
    \end{tcolorbox}
    \caption{Step-by-step solution prompt. Directly replicates original instructions for mathematical problem solving with explicit thinking requirement.}
    \label{fig:multimodal_eval_zero_shot_cot_prompt}
\end{figure*}
\begin{figure*}[h]
    \centering
    \tcbset{colframe=black, colback=gray!10, arc=5mm}
    \begin{tcolorbox}
    \small
    \textbf{Your role as an assistant involves thoroughly exploring questions through a systematic long thinking process before providing the final precise and accurate solutions. This requires engaging in a comprehensive cycle of analysis, summarizing, exploration, reassessment, reflection, backtracing, and iteration to develop well-considered thinking process.}

    \textbf{Please structure your response into two main sections: Thought and Solution.}

    \textbf{In the Thought section, detail your reasoning process using the specified format:}
    <|begin\_of\_thought|> \{\{thought with steps separated with '\textbackslash n\textbackslash n'\}\} 
    <|end\_of\_thought|>

    \textbf{Each step should include detailed considerations such as:}
    - Analyzing questions
    - Summarizing relevant findings
    - Brainstorming new ideas
    - Verifying the accuracy of current steps
    - Refining any errors
    - Revisiting previous steps

    \textbf{In the Solution section, present the final solution that you deem correct:}
    <|begin\_of\_solution|> 
    \{\{final formatted, precise, and clear solution\}\} 
    <|end\_of\_solution|>

    \textbf{Requirements:}
    - Put your final answer within \textbackslash boxed\{\{answer\}\}
    - Maintain logical, accurate, and concise expression
    - Include all necessary steps to reach the conclusion

    \textbf{Now, try to solve the following question through the above guidelines:}

    <image> \{question\}
    \end{tcolorbox}
    \caption{Structured problem-solving prompt. Specifies comprehensive thinking process with separate Thought and Solution sections, requiring detailed reasoning before final answer.}
    \label{fig:multimodal_eval_long_cot_prompt}
\end{figure*}

\subsection{LLM-as-Judge Evaluation}
We provide our prompts for clarity, completeness, correction, and redundancy in Figures~\ref{fig:clarity_eval_prompt},~\ref{fig:completeness_eval_prompt},~\ref{fig:correctness_eval_prompt},~\ref{fig:redundancy_eval_prompt}.

\begin{figure*}[h!]
    \centering
    \begin{subfigure}[b]{0.5\textwidth}
        \includegraphics[width=\linewidth]{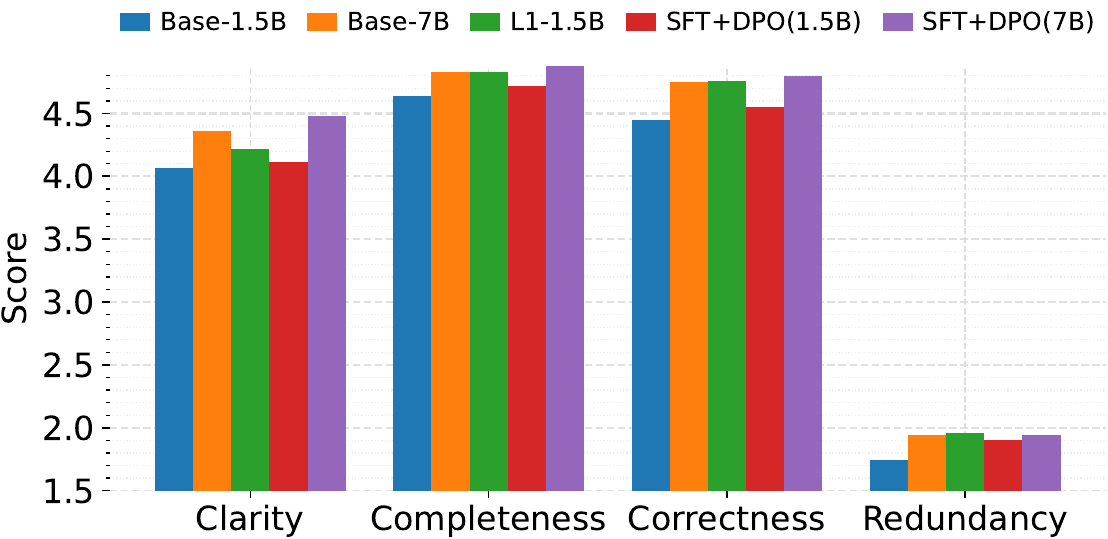}
        \caption{Unimodal Scores.}
        \label{fig:unimodel_cotscores}
    \end{subfigure}
    \hspace{-0.2cm}
    \begin{subfigure}[b]{0.5\textwidth}
        \includegraphics[width=\linewidth]{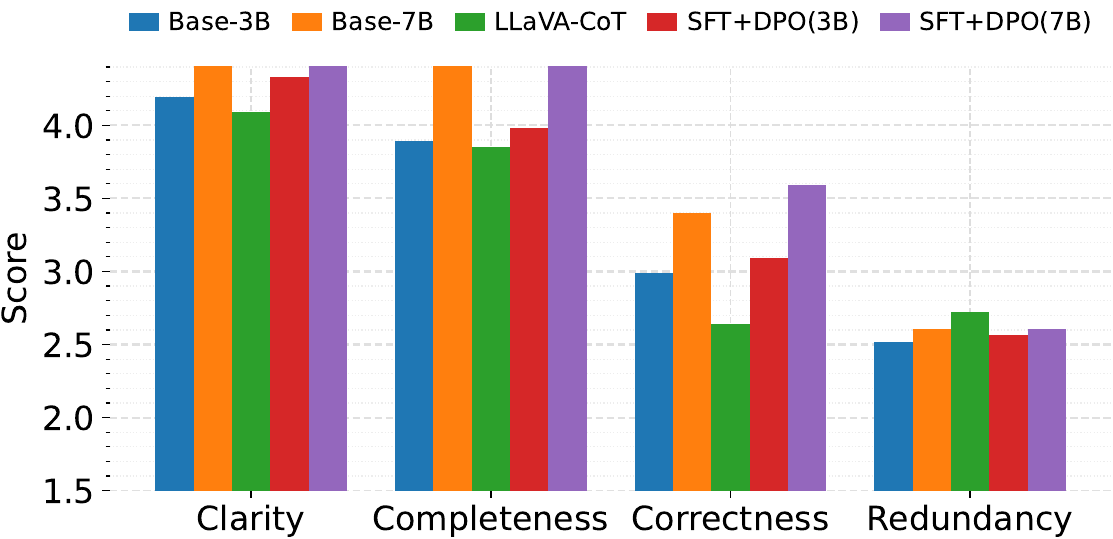}
        \caption{Multimodal Scores.}
        \label{fig:multimodal_cotscores}
    \end{subfigure}
    \caption{LLM-as-Judge evaluation scores with chain-of-thought rationale.}
    \label{fig:llm_as_judge_scores}
\end{figure*}

\begin{figure*}[h]
    \centering
    \tcbset{colframe=black, colback=gray!10, arc=5mm}
    \begin{tcolorbox}
    \small
    \textbf{You will be given a solution for a math problem. Your task is to evaluate the solution for its clarity.} \\

    \textbf{Evaluation Criteria:}

    \textbf{Clarity (1–5):} Evaluates how clearly the solution communicates its reasoning.

    \begin{itemize}
        \item \textbf{Score: 5} – Exceptionally clear with perfect logical flow and insightful explanations.
        \item \textbf{Score: 4} – Clear and well-structured with good explanations.
        \item \textbf{Score: 3} – Somewhat clear but could benefit from better organization.
        \item \textbf{Score: 2} – Poorly structured with unclear connections between steps.
        \item \textbf{Score: 1} – Confusing and disorganized, very difficult to follow.
    \end{itemize}

    \textbf{Evaluation Steps:}
    \begin{enumerate}
        \item Carefully read both the problem and the provided solution.
        \item Examine the logical progression from one step to the next.
        \item Determine whether key concepts and reasoning steps are explained in a clear and concise manner.
        \item Based on the criteria above, assign a clarity score from 1 to 5.
    \end{enumerate}

    \textbf{Instructions:}
    \begin{itemize}
        \item \textbf{Do not} attempt to solve, fix, or fact-check the solution.
        \item Focus solely on how clearly the reasoning is communicated.
        \item Provide your score in the following format: \texttt{[[score]]}
    \end{itemize}

    \textbf{Input:}  \\
    \textbf{Problem:} \{question\}  \\
    \textbf{Solution:} \{solution\} \\

    \textbf{Output (score only):}  \\
    \texttt{- Clarity: [[score]]}
    \end{tcolorbox}
    \caption{Prompt for evaluating clarity of mathematical solutions.}
    \label{fig:clarity_eval_prompt}
\end{figure*}

\begin{figure*}[h]
    \centering
    \tcbset{colframe=black, colback=gray!10, arc=5mm}
    \begin{tcolorbox}
    \small
    \textbf{You will be given a solution for a math problem. Your task is to evaluate the solution for its completeness.} \\

    \textbf{Evaluation Criteria:}

    \textbf{Completeness (1–5):} Evaluates whether all essential reasoning steps are included.

    \begin{itemize}
        \item \textbf{Score: 5} – Perfectly complete, including all necessary steps while maintaining conciseness.
        \item \textbf{Score: 4} – Contains all essential reasoning steps with only minor details omitted.
        \item \textbf{Score: 3} – Most essential steps are included, but some connections require inference.
        \item \textbf{Score: 2} – Several important steps are missing, creating significant gaps.
        \item \textbf{Score: 1} – Critical reasoning steps are missing, making it impossible to follow.
    \end{itemize}

    \textbf{Evaluation Steps:}
    \begin{enumerate}
        \item Carefully read both the problem and the provided solution.
        \item Identify the key logical and mathematical steps required to solve the problem.
        \item Assess whether these steps are present and clearly stated in the solution.
        \item Based on the criteria above, assign a completeness score from 1 to 5.
    \end{enumerate}

    \textbf{Instructions:}
    \begin{itemize}
        \item \textbf{Do not} attempt to solve, fix, or fact-check the solution.
        \item Focus solely on whether all essential reasoning steps are present.
        \item Provide your score in the following format: \texttt{[[score]]}
    \end{itemize}

    \textbf{Input:}  \\
    \textbf{Problem:} \{question\}  \\
    \textbf{Solution:} \{solution\} \\

    \textbf{Output (score only):}  \\
    \texttt{- Completeness: [[score]]}
    \end{tcolorbox}
    \caption{Prompt for evaluating completeness of mathematical solutions.}
    \label{fig:completeness_eval_prompt}
\end{figure*}

\begin{figure*}[h]
    \centering
    \tcbset{colframe=black, colback=gray!10, arc=5mm}
    \begin{tcolorbox}
    \small
    \textbf{You will be given a solution for a math problem. Your task is to evaluate the solution for the correctness of its intermediate steps.} \\

    \textbf{Evaluation Criteria:}

    \textbf{Correctness of Intermediate Steps (1–5):} Evaluates the mathematical accuracy of each intermediate reasoning step, not just the final answer.

    \begin{itemize}
        \item \textbf{Score: 5} – All intermediate reasoning steps are mathematically correct and logically sound.
        \item \textbf{Score: 4} – Nearly all steps are mathematically correct with only very minor errors.
        \item \textbf{Score: 3} – The majority of steps are mathematically sound, but there are a few minor errors.
        \item \textbf{Score: 2} – Some steps are mathematically incorrect or invalid logical connections.
        \item \textbf{Score: 1} – Most intermediate steps contain significant mathematical errors or invalid logical connections.
    \end{itemize}

    \textbf{Evaluation Steps:}
    \begin{enumerate}
        \item Carefully read both the problem and the provided solution.
        \item Identify and examine each intermediate reasoning step.
        \item Determine whether each step is mathematically correct and logically valid.
        \item Based on the criteria above, assign a correctness score from 1 to 5.
    \end{enumerate}

    \textbf{Instructions:}
    \begin{itemize}
        \item \textbf{Do not} solve or complete the problem yourself.
        \item Focus only on evaluating the accuracy of the intermediate steps shown in the solution.
        \item Provide your score in the following format: \texttt{[[score]]}
    \end{itemize}

    \textbf{Input:}  \\
    \textbf{Problem:} \{question\}  \\
    \textbf{Solution:} \{solution\} \\

    \textbf{Output (score only):}  \\
    \texttt{- Correctness of Intermediate Steps: [[score]]}
    \end{tcolorbox}
    \caption{Prompt for evaluating correctness of intermediate steps in mathematical solutions.}
    \label{fig:correctness_eval_prompt}
\end{figure*}

\begin{figure*}[h]
    \centering
    \tcbset{colframe=black, colback=gray!10, arc=5mm}
    \begin{tcolorbox}
    \small
    \textbf{You will be given a solution for a math problem. Your task is to evaluate the solution for its redundancy.} \\

    \textbf{Evaluation Criteria:}

    \textbf{Redundancy (1–5):} Evaluates the presence of unnecessary repetition or redundant steps in the solution.

    \begin{itemize}
        \item \textbf{Score: 5} – Contains no redundancy – each step builds on previous ones without unnecessary repetition.
        \item \textbf{Score: 4} – Mostly free of redundancy, with minimal unnecessary repetition.
        \item \textbf{Score: 3} – Contains some redundancy but most repetition serves a purpose.
        \item \textbf{Score: 2} – Shows significant redundancy with several instances of unnecessary repetition.
        \item \textbf{Score: 1} – Contains extensive redundancy with multiple unnecessary repetitions.
    \end{itemize}

    \textbf{Evaluation Steps:}
    \begin{enumerate}
        \item Carefully read both the problem and the provided solution.
        \item Identify any repetition or restatement of the same concepts or steps.
        \item Assess whether the repetition is necessary for clarity or if it is excessive and avoidable.
        \item Based on the criteria above, assign a redundancy score from 1 to 5.
    \end{enumerate}

    \textbf{Instructions:}
    \begin{itemize}
        \item \textbf{Do not} attempt to solve or correct the solution.
        \item Focus only on identifying and evaluating unnecessary repetition or redundant reasoning.
        \item Provide your score in the following format: \texttt{[[score]]}
    \end{itemize}

    \textbf{Input:}  \\
    \textbf{Problem:} \{question\}  \\
    \textbf{Solution:} \{solution\} \\

    \textbf{Output (score only):}  \\
    \texttt{- Redundancy: [[score]]}
    \end{tcolorbox}
    \caption{Prompt for evaluating redundancy in mathematical solutions.}
    \label{fig:redundancy_eval_prompt}
\end{figure*}

\begin{figure*}[h]
    \centering
    \tcbset{colframe=black, colback=gray!10, arc=5mm}
    \begin{tcolorbox}
    \small
    % --- Unchanged content from the original prompt is omitted ---

    % NEW/UPDATED SECTIONS ONLY:

    \textbf{Instructions:}
    \begin{itemize}
        \item \textbf{Do not} attempt to solve, fix, or fact-check the solution.
        \item Focus only on evaluating how clearly the reasoning is communicated.
        \item First, provide your reasoning inside \texttt{<think>...</think>}.
        \item Then, output the final clarity score in the format \texttt{<score>[[score]]</score>}.
    \end{itemize}

    \textbf{Output:}  \\
    \texttt{<think>} \\
    \texttt{[Your detailed step-by-step reasoning about the clarity of the solution and justification for the score goes here.]} \\
    \texttt{</think>} \\[0.5em]
    \texttt{<score>[[score]]</score>}
    \end{tcolorbox}
    \caption{Extended clarity evaluation prompt with rationale and explicit XML-style output format.}
    \label{fig:clarity_eval_extended}
\end{figure*}

\subsection{Results}~\label{appsubsec:results}

\begin{figure*}[h!]
    \centering
    \begin{subfigure}[b]{0.30\textwidth}
        \includegraphics[width=\linewidth]{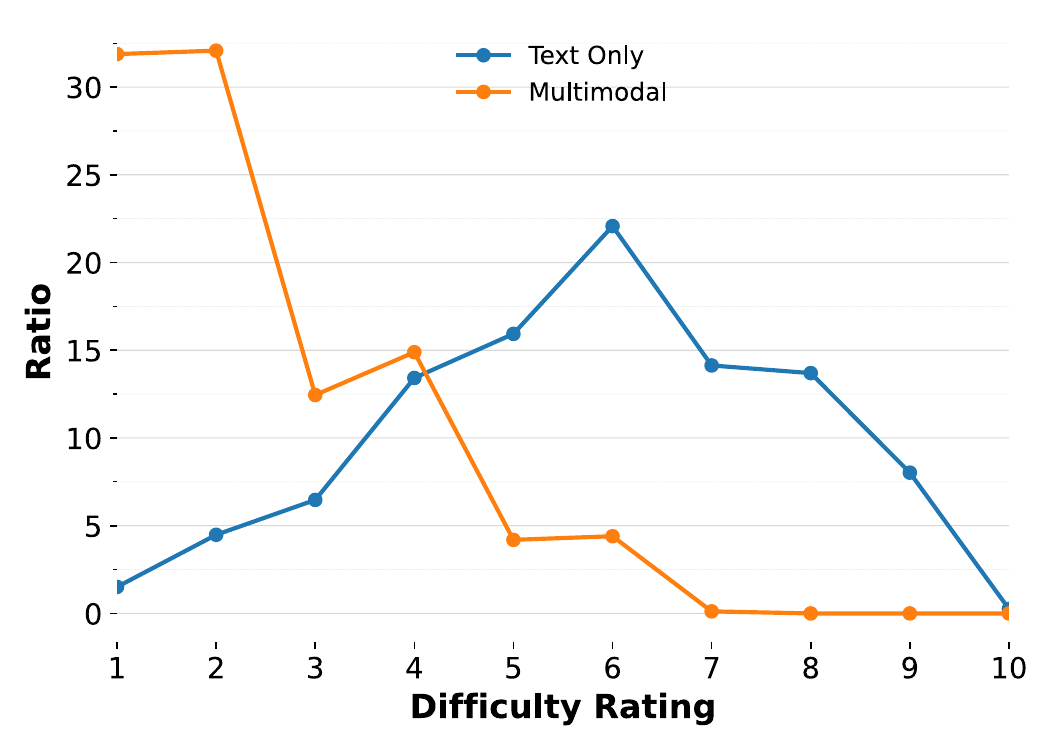}
        \caption{Difficulty distribution in training data.}
        \label{fig:train_ratings}
    \end{subfigure}
    \hfill
    \begin{subfigure}[b]{0.30\textwidth}
        \includegraphics[width=\linewidth]{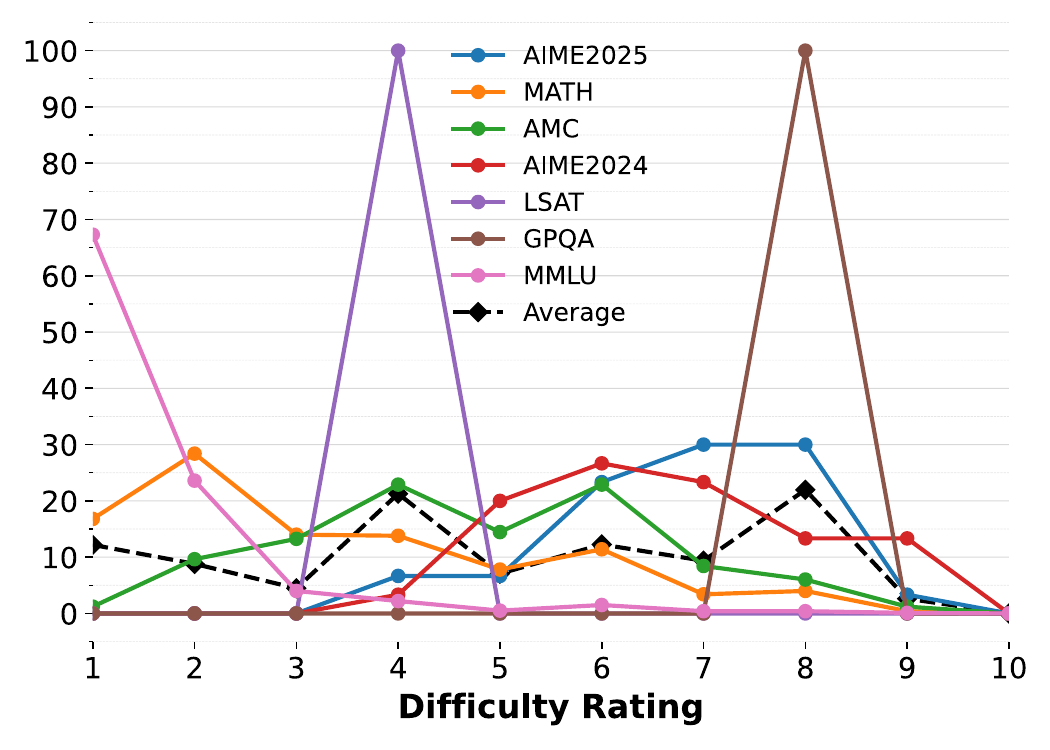}
        \caption{Difficulty distribution in text-only evaluation benchmarks.}
        \label{fig:text_only_ratings}
    \end{subfigure}
    \hfill
    \begin{subfigure}[b]{0.30\textwidth}
        \includegraphics[width=\linewidth]{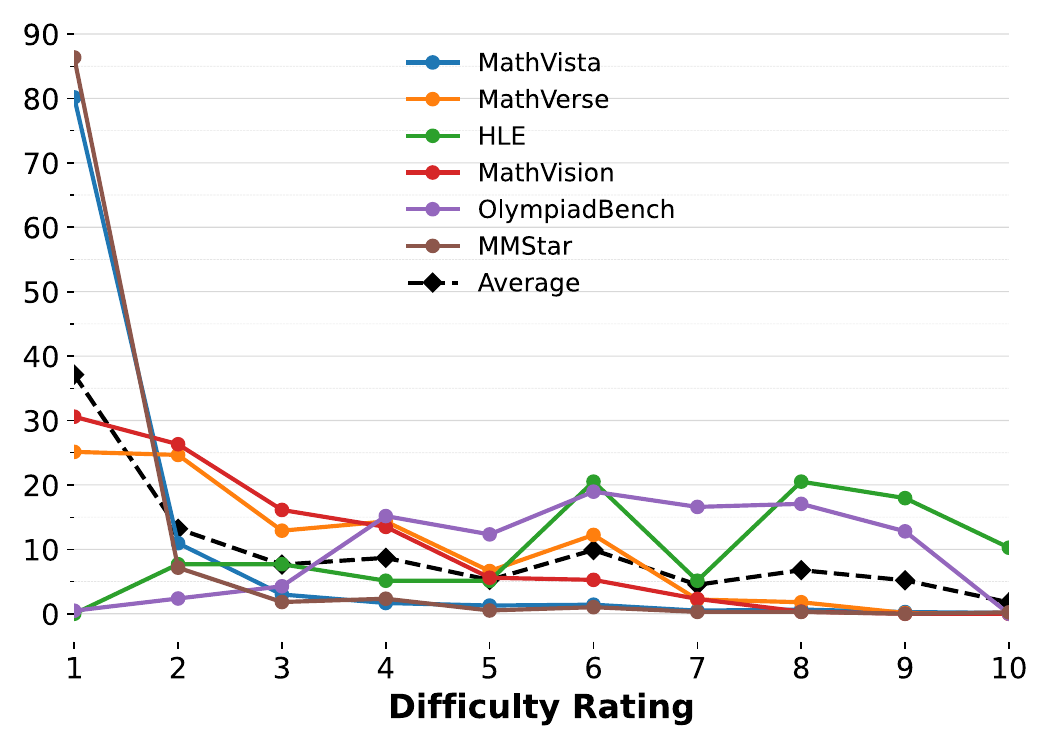}
        \caption{Difficulty distribution in multimodal evaluation datasets.}
        \label{fig:multimodal_ratings}
    \end{subfigure}
    \caption{
Difficulty ratings assigned to each problem using the AoPS scale. 
\textbf{(a)} Distribution of difficulty scores for the training dataset. 
\textbf{(b)} Distribution for text-only evaluation benchmarks, which skew toward higher difficulty levels. 
\textbf{(c)} Multimodal evaluation datasets tend to contain easier problems overall. These ratings guide our compression process, enabling adaptive reasoning trace lengths based on problem complexity.
}
\label{fig:difficulty_ratings}
\end{figure*}

In this section, we present the additional results to support our work. 

\subsection{Evaluation of Summarization Approaches for CoT Compression}

To develop an effective difficulty-aware reasoning framework, we first investigated different summarization techniques to compress reasoning trajectories. Our objective was to identify methods that maximize compression while preserving essential reasoning information across problems of varying difficulty.

We compared six distinct approaches to generate and summarize chains of thought. As shown in Table~\ref{tab:summarization}, these approaches yielded significantly different compression ratios when applied to the same set of mathematical problems.

Our baseline \texttt{long\_cot} consists of full reasoning traces generated by DeepSeek-R1, representing the verbose reasoning typically produced by advanced language models. For comparison, we evaluated a single-step approach (\texttt{succinct\_cot}) where we directly instructed GPT-4o to produce concise reasoning.

We then explored various two-step summarization approaches using GPT-4o to compress the original DeepSeek-R1 reasoning:

\begin{itemize}
    \item The \texttt{gpt4o\_basic} approach applied a general summarization prompt without difficulty awareness, achieving the highest compression ratio at 79.1\% (reducing from 4702.3 to 982.8 tokens on average).
    \item The \texttt{gpt4o\_num} variant enforced a structured output format with numbered reasoning steps, resulting in a 55.2\% reduction.
    \item Finally, we tested two difficulty-aware variants: \texttt{gpt4o\_DA1} incorporated difficulty ratings to guide compression level, while \texttt{gpt4o\_DA2} combined difficulty awareness with the structured output format.
\end{itemize}

\begin{table}[h!]
    \centering
    \begin{tabular}{lcc}
        \toprule
        \textbf{Setup} & \textbf{Avg Tokens} & \textbf{Max Reduction} \\
        \midrule
        \textcolor{gray}{long\_cot} & \textcolor{gray}{4702.3} & \textcolor{gray}{0.0\%} \\
        succinct\_cot & 3456.2 & 26.5\% \\
        gpt4o\_basic & \textbf{982.8} & \textbf{79.1\%} \\
        gpt4o\_num & 2106.6 & 55.2\% \\
        gpt4o\_DA1 & 2473.4 & 47.4\% \\
        gpt4o\_DA2 & \underline{2036.1} & \underline{56.7\%} \\
        \bottomrule
    \end{tabular}
    \caption{Summarization results and compression ratio with different prompting approaches.}
    \label{tab:summarization}
\end{table}

The difficulty-aware approach with structured output format (\texttt{gpt4o\_DA2}) achieved an optimal balance, reducing token usage by 56.7\% while maintaining appropriate reasoning detail proportional to problem complexity. This aligns with our goal of teaching models to ``think hard'' on difficult problems while being concise on simpler ones.

While \texttt{gpt4o\_basic} achieved the highest compression, our qualitative analysis revealed that it often omitted critical reasoning steps needed for more difficult problems. In contrast, \texttt{gpt4o\_DA2} produced summaries that appropriately scaled with problem difficulty—remaining concise for simpler problems while preserving necessary detail for complex ones. Based on these findings, we selected \texttt{gpt4o\_DA2} as our primary summarization approach for generating training data in subsequent experiments.

\subsection{Error Analysis}\label{appsubsec:error_analysis}
\begin{table*}[ht]
% \centering
% \renewcommand{\arraystretch}{1.5}
% \setlength{\tabcolsep}{10pt}
% \setlength{\tabcolsep}{6pt}
% \renewcommand{\arraystretch}{1.2}XX

\centering
\small
\setlength{\tabcolsep}{4pt}  % tweak column spacing if needed
\renewcommand{\arraystretch}{0.9}  % tighten row spacing

\resizebox{\linewidth}{!}{%

\begin{tabular}{>{\centering\arraybackslash}m{0.25\textwidth} p{0.33\textwidth} p{0.33\textwidth}}
% \begin{tabular}{ccc}
\toprule
\textbf{Example} & \textbf{Qwen-2.5-VL-7B (SFT + DPO)} & \textbf{LLava-CoT} \\
\midrule

\parbox[c]{\linewidth}{
    \includegraphics[width=0.9\linewidth]{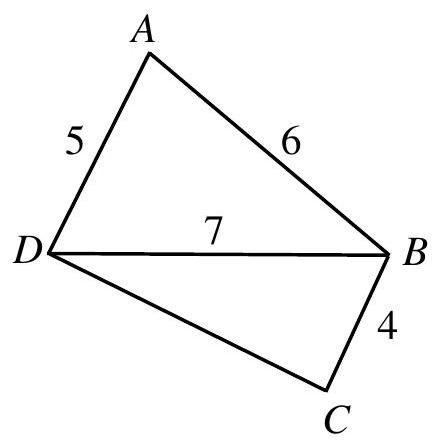}\\
    % \vspace{2pt}
    \textit{In the diagram, $A B C D$ is a quadrilateral in which $\angle A+\angle C=180^{\circ}$. What is the length of $C D$ ?}
} &

To find \( CD \), we are given: \( AB = 6 \), \( AD = 5 \), \( BC = 4 \), \( BD = 7 \), and \( \angle A + \angle C = 180^\circ \). Since opposite angles are supplementary, quadrilateral \( ABCD \) is cyclic.

\textbf{Triangle \( ABD \):}  
Apply the Law of Cosines:
\[
49 = 36 + 25 - 60\cos(\angle A) \Rightarrow \cos(\angle A) = \frac{1}{5}
\]

\textbf{Triangle \( BCD \):}  
Using \( \cos(\angle C) = -\cos(\angle A) \), apply the Law of Cosines:
\[
49 = 16 + x^2 + \frac{8x}{5} \Rightarrow 5x^2 + 8x - 165 = 0
\]

Solving the quadratic:
\[
x = \frac{-8 \pm \sqrt{3364}}{10} = \frac{-8 \pm 58}{10} \Rightarrow x = 5
\]

\textbf{Final Answer:} \( \boxed{5} \) &

Given \( AB = 6 \), \( AD = 5 \), \( BC = 4 \), and \( \angle A + \angle C = 180^\circ \), the reasoning incorrectly assumes that \( ABCD \) is a parallelogram.

It then wrongly applies the property that opposite sides are equal:
\[
CD = AB = 6
\]

\textbf{Final Answer:} \( \boxed{6} \) \hfill (\textcolor{red}{Incorrect}) \\

\midrule

\parbox[c]{\linewidth}{
    \includegraphics[width=0.9\linewidth]{assets/multimodal_p2.png}\\
    % \vspace{2pt}
    \textit{}
} &

To determine the discontinuities of \( f(x) \), we analyze the graph step-by-step:

\begin{itemize}
    \item \textbf{At \( x = 0 \)}:  
    Left-hand limit = 2, right-hand limit = 1. Since these limits differ, there is a \textbf{jump discontinuity}.
    
    \item \textbf{At \( x = 1 \)}:  
    Left-hand limit = 1, right-hand limit = 2. Again, a \textbf{jump discontinuity}.
    
    \item \textbf{At \( x = 2 \)}:  
    Both left and right limits = 2, and \( f(2) = 2 \). Hence, the function is \textbf{continuous}.
    
    \item \textbf{At \( x = 3 \)}:  
    Left and right limits = 1, and \( f(3) = 1 \). Therefore, the function is \textbf{continuous}.
\end{itemize}

\textbf{Discontinuities:} \( x = 0 \) and \( x = 1 \)

\textbf{Final Answer:} \( \boxed{A} \) &

The solution examines the graph and claims vertical asymptotes at:

\begin{itemize}
    \item \( x = 0 \)
    \item \( x = 2 \)
\end{itemize}

It assumes these are discontinuities where the function is undefined and compares to options:

\begin{itemize}
    \item (A) \( x = 0 \), \( x = 1 \)
    \item (D) \( x = 2 \)
\end{itemize}

The reasoning concludes that both (A) and (D) seem correct and chooses:

\textbf{Final Answer:} \( \boxed{A} \)

\\

\bottomrule
\end{tabular}
}
\caption{Qualitative comparison of reasoning traces from our Qwen-2.5-VL-7B (SFT + DPO) model and the LLaVA-CoT baseline on two multimodal math problems. Each example includes the input image and question, followed by the model's reasoning trace and final answer. In the first example, our model correctly identifies the geometric structure and applies the Law of Cosines to compute the answer, while LLaVA-CoT incorrectly assumes a parallelogram and reaches an incorrect conclusion. In the second example, our model accurately identifies jump discontinuities based on left and right limits, whereas LLaVA-CoT misinterprets vertical asymptotes and includes a spurious discontinuity. These examples illustrate how difficulty-aware, compressed reasoning enables more accurate and structured outputs compared to baseline multimodal CoT models.}
\label{tab:multimodal_errors}

\end{table*}

In this section, we provide chain-of-thought reasoning traces generated by different models across unimodal and multimodal setups.

\subsubsection{Unimodal}
In Table~\ref{tab:unimodel_errors} we show quality chain-of-thought examples generated by unimodal models.
\subsubsection{Multimodal}
In Table~\ref{tab:multimodal_errors}, we show a few qualitative examples of the chain-of-thought reasoning process (annotated) by our model (7B SFT + DPO) and LLava-CoT baseline.

\begin{table*}[ht]
\centering
\small
\setlength{\tabcolsep}{4pt}
\renewcommand{\arraystretch}{0.9}
\resizebox{\linewidth}{!}{%
\begin{tabular}{>{\raggedright\arraybackslash}p{0.25\textwidth} >{\raggedright\arraybackslash}p{0.33\textwidth} >{\raggedright\arraybackslash}p{0.33\textwidth}}
\toprule
\textbf{Question} & \textbf{R1-Distill-Qwen-7B (SFT+DPO)} & \textbf{R1-Distill-Qwen-7B} \\
\midrule
Let $\mathbf{A}$ be a matrix such that
$\mathbf{A} \begin{pmatrix} 3 \\ 1 \\ 0 \end{pmatrix} = \begin{pmatrix} 2 \\ 4 \\ -3 \end{pmatrix}$ and
$\mathbf{A} \begin{pmatrix} -5 \\ 2 \\ 2 \end{pmatrix} = \begin{pmatrix} 0 \\ -5 \\ -5 \end{pmatrix}$.

Find $\mathbf{A} \begin{pmatrix} -13 \\ 3 \\ 4 \end{pmatrix}$.
& 
Step 1: Let $\mathbf{v}_1 = \begin{pmatrix} 3 \\ 1 \\ 0 \end{pmatrix}$,
$\mathbf{v}_2 = \begin{pmatrix} -5 \\ 2 \\ 2 \end{pmatrix}$, and express: 
$\begin{pmatrix} -13 \\ 3 \\ 4 \end{pmatrix} = c_1 \mathbf{v}_1 + c_2 \mathbf{v}_2$

Step 2: Solve system:
$2c_2 = 4 \Rightarrow c_2 = 2$
$c_1 + 2c_2 = 3 \Rightarrow c_1 = -1$
Confirm: $3c_1 - 5c_2 = -13$

Step 3: Use linearity: 
$\mathbf{A}x = -1 \cdot \mathbf{A}v_1 + 2 \cdot \mathbf{A}v_2$

Step 4: Compute:
$-1 \cdot \begin{pmatrix} 2 \\ 4 \\ -3 \end{pmatrix} + 2 \cdot \begin{pmatrix} 0 \\ -5 \\ -5 \end{pmatrix} = \boxed{\begin{pmatrix} -2 \\ -14 \\ -7 \end{pmatrix}}$
& 
Step 1: Express $\mathbf{A}$ via outer products:
$\mathbf{A} = a \mathbf{v}_1 \mathbf{w}_1^T + b \mathbf{v}_2 \mathbf{w}_2^T$

Step 2: Dot products:
$w_1^T x = -26$, $w_2^T x = -35$

Step 3: Compute:
$\mathbf{A}x = -26 \cdot \mathbf{v}_1 + (-35) \cdot \mathbf{v}_2 = \begin{pmatrix} 97 \\ -96 \\ -70 \end{pmatrix}$

Step 4: Incorrect final result: $\boxed{\begin{pmatrix} 97 \\ -96 \\ -70 \end{pmatrix}}$ \\
\midrule
Let \( P(x) \) be a polynomial with rational coefficients such that when \( P(x) \) is divided by \( x^2 + x + 1 \), the remainder is \( x + 2 \), and when divided by \( x^2 + 1 \), the remainder is \( 2x + 1 \).

What is the sum of the squares of the coefficients of the polynomial of least degree with these properties?
& 
Step 1: Use polynomial division identity:
\( P(x) = (x^2 + x + 1)Q_1(x) + (x + 2) \)
\( P(x) = (x^2 + 1)Q_2(x) + (2x + 1) \)

Step 2: Set equal and subtract:
\((x^2 + x + 1)Q_1(x) - (x^2 + 1)Q_2(x) = x - 1\)

Step 3: Let \( Q_1 = ax + b, Q_2 = cx + d \), solve:
\( a = 1, b = 1, c = 1, d = 2 \)
So \( P(x) = (x^2 + x + 1)(x + 1) + (x + 2) = x^3 + 2x^2 + 3x + 3 \)

Step 4: Try lower degree → contradiction

Step 5: Sum of squares: \( 1^2 + 2^2 + 3^2 + 3^2 = \boxed{23} \)
& 
Step 1: Recognizes equivalence via polynomial division

Step 2: Uses Chinese Remainder Theorem logic:
\( P(x) = (x^2 + x + 1)^2 U(x) + x^2 + 2x + 3 \)

Step 3: Set \( U(x) = 1 \)
So \( P(x) = (x^2 + x + 1)^2 + x^2 + 2x + 3 \)
Compute: \( x^4 + 2x^3 + 4x^2 + 4x + 4 \)

Step 4: Sum of squares: \( 1^2 + 2^2 + 4^2 + 4^2 + 4^2 = \boxed{53} \) \\
\midrule
Evaluate
$\sin (\arcsin 0.4 + \arcsin 0.5) \cdot \sin (\arcsin 0.5 - \arcsin 0.4)$
& 
Step 1: Let \( A = \arcsin 0.5 \), \( B = \arcsin 0.4 \)
Then the expression becomes \( \sin(A + B) \cdot \sin(A - B) \)

Step 2: Use identity:
\( \sin(A + B)\sin(A - B) = \sin^2 A - \sin^2 B \)

Step 3: \( \sin^2 A = 0.25 \), \( \sin^2 B = 0.16 \)

Step 4: Difference: \( 0.25 - 0.16 = \boxed{0.09} \)
& 
Step 1: Same substitution: \( A = \arcsin 0.5 \), \( B = \arcsin 0.4 \)

Step 2: Applies identity but subtracts wrong:
\( \sin^2 A - \sin^2 B = 0.16 - 0.25 = \boxed{-0.09} \) \\
\bottomrule
\end{tabular}
}
\caption{Qualitative comparison of reasoning traces from our R1-Distill-Qwen-7B (SFT+DPO) model and its base counterpart on three math problems: matrix transformation, polynomial reconstruction, and trigonometric identity evaluation. In each case, the SFT+DPO model demonstrates structured, accurate reasoning using appropriate mathematical identities and problem decomposition. In contrast, the base model often makes conceptual or arithmetic errors—such as misapplying polynomial congruences or miscomputing identity results—leading to incorrect final answers. These examples illustrate the improved reasoning quality and reliability gained through difficulty-aware distillation.}
\label{tab:unimodel_errors}
\end{table*}

% \subsection{Hyperparameters}\label{appsubsec:hyperparameters}
% \input{tables/hyperparameters}

\end{document}